%% file: main.tex
\documentclass[ twoside,openright,titlepage,numbers=noenddot,
                headinclude,footinclude,cleardoublepage=empty,abstract=on,
                BCOR=5mm,paper=a4,fontsize=11pt
                ]{scrreprt}

\input{classicthesis-config}

\usepackage{todonotes} %
\usepackage{siunitx} %
\usepackage{amssymb} %
\usepackage{tcolorbox} %
\usepackage{bm}

\begin{document}
\frenchspacing
\raggedbottom
\selectlanguage{american} 
\setbibpreamble{}
\pagenumbering{roman}
\pagestyle{plain}
\newcommand\eqdef{\ \mathrel{\overset{\makebox[0pt]{\mbox{\normalfont\scriptsize\sffamily def}}}{=}}\ }

\include{FrontBackmatter/Titlepage}
\include{FrontBackmatter/Titleback}


\cleardoublepage\include{FrontBackmatter/Abstract}
\cleardoublepage\include{FrontBackmatter/Contents}

\cleardoublepage
\pagestyle{scrheadings}
\pagenumbering{arabic}
\cleardoublepage

\include{Chapters/Chapter01}
\include{Chapters/Chapter02}

\include{Chapters/Chapter03}
\include{Chapters/Chapter04}
\include{Chapters/Chapter06}

\appendix
\cleardoublepage
\include{Chapters/Chapter0A}

\cleardoublepage\include{FrontBackmatter/Bibliography}
\end{document}

%% file: classicthesis-config.tex
\PassOptionsToPackage{utf8}{inputenc}
  \usepackage{inputenc}

\PassOptionsToPackage{T1}{fontenc} 
  \usepackage{fontenc}

\PassOptionsToPackage{
  drafting=false,    
  tocaligned=false, 
  dottedtoc=false,  
  eulerchapternumbers=true, 
  linedheaders=false,       
  floatperchapter=true,     
  eulermath=false,  
  beramono=false,    
  palatino=false,    
  style=classicthesis 
}{classicthesis}

\newcommand{\myTitle}{Including STDP to eligibility propagation in multi-layer recurrent spiking neural networks\xspace}
\newcommand{\mySubtitle}{Master's Thesis\xspace}
\newcommand{\myDegree}{MSc. Artificial Intelligence\xspace}
\newcommand{\myName}{Werner van der Veen\xspace}

\newcommand{\myFaculty}{Faculty of Science and Engineering\xspace}

\newcommand{\myUni}{University of Groningen\xspace}

\newcommand{\myTime}{May 3, 2021\xspace}
\newcommand{\myVersion}{Put version here}

\providecommand{\mLyX}{L\kern-.1667em\lower.25em\hbox{Y}\kern-.125emX\@}
\newcommand{\ie}{i.\,e.}

\newcommand{\eg}{e.\,g.}

\PassOptionsToPackage{american}{babel} 
\usepackage{babel}

\usepackage{csquotes}
\PassOptionsToPackage{%
  backend=biber,bibencoding=utf8, 
  language=auto,%
  style=authoryear-comp,%
  citestyle=authoryear-comp, 
  bibstyle=authoryear,dashed=false, 
  sorting=nyt, 
  maxbibnames=3, 
  natbib=true 
}{biblatex}
    \usepackage{biblatex}

\PassOptionsToPackage{fleqn}{amsmath}       
  \usepackage{amsmath}

\usepackage{graphicx} %
\usepackage{scrhack} 
\usepackage{xspace} 
\PassOptionsToPackage{printonlyused,smaller}{acronym}
  \usepackage{acronym} 

\usepackage{tabularx} 
\newcommand{\tableheadline}[1]{\multicolumn{1}{l}{\spacedlowsmallcaps{#1}}}
\newcommand{\myfloatalign}{\centering} 
\usepackage{subfig}

\usepackage{listings}
\usepackage{classicthesis}

\hypersetup{%
  colorlinks=false, linktocpage=true, pdfstartpage=3, pdfstartview=FitV,%
  breaklinks=true, pageanchor=true,%
  pdfpagemode=UseNone, %
  plainpages=false, bookmarksnumbered, bookmarksopen=true, bookmarksopenlevel=1,%
  hypertexnames=true, pdfhighlight=/O,
  urlcolor=CTurl, linkcolor=CTlink, citecolor=CTcitation, 
  pdftitle={\myTitle},%
  pdfauthor={\textcopyright\ \myName, \myUni, \myFaculty},%
  pdfsubject={},%
  pdfkeywords={},%
  pdfcreator={pdfLaTeX},%
  pdfproducer={LaTeX with hyperref and classicthesis}%
}

\makeatletter
\@ifpackageloaded{babel}%
  {%
    \addto\extrasamerican{%
    }%
    }{\relax}
\makeatother

%% file: FrontBackmatter/Titlepage.tex
\begin{titlepage}
    \begin{addmargin}[-1cm]{-3cm}
    \begin{center}
        \includegraphics[width=10cm]{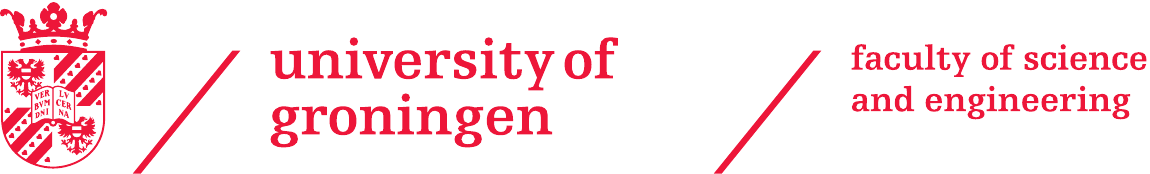} \\ \medskip
        \large

        \hfill

        \vfill

        \begingroup
            \color{CTtitle}\spacedallcaps{\myTitle} \\ \bigskip
        \endgroup

        \spacedlowsmallcaps{\myName}

        \vfill

        \myDegree \\
        \myFaculty \\
        \myUni \\ \bigskip
        Supervised by Dr. Herbert Jaeger \& Dr. Marco Wiering \\ \bigskip

        \myTime

        \vfill

    \end{center}
  \end{addmargin}
\end{titlepage}

%% file: FrontBackmatter/Titleback.tex
\thispagestyle{empty}

\hfill

\vfill

\noindent\myName: \textit{\myTitle,} \mySubtitle, 
\textcopyright\ \myTime

%
%
%
%
%

%% file: FrontBackmatter/Abstract.tex
\pdfbookmark[1]{Abstract}{Abstract}
\begingroup
\let\clearpage\relax
\let\cleardoublepage\relax
\let\cleardoublepage\relax

\chapter*{Abstract}
Spiking neural networks (SNNs) in neuromorphic systems are more energy efficient compared to deep learning--based methods, but there is no clear competitive learning algorithm for training such SNNs.
Eligibility propagation (e-prop) offers an efficient and biologically plausible way to train competitive recurrent SNNs in low-power neuromorphic hardware.
In this report, previous performance of e-prop on a speech classification task is reproduced, and the effects of including STDP-like behavior are analyzed.
Including STDP to the ALIF neuron model improves the classification performance, but this is not the case for the Izhikevich e-prop neuron.
Finally, it was found that e-prop implemented in a single-layer recurrent SNN consistently outperforms a multi-layer variant.

\endgroup

\vfill

%% file: FrontBackmatter/Contents.tex
\pagestyle{scrheadings}
\pdfbookmark[1]{\contentsname}{tableofcontents}
\setcounter{tocdepth}{2} 
\setcounter{secnumdepth}{3} 
\manualmark
\markboth{\spacedlowsmallcaps{\contentsname}}{\spacedlowsmallcaps{\contentsname}}
\tableofcontents
\automark[section]{chapter}
\renewcommand{\chaptermark}[1]{\markboth{\spacedlowsmallcaps{#1}}{\spacedlowsmallcaps{#1}}}
\renewcommand{\sectionmark}[1]{\markright{\textsc{\thesection}\enspace\spacedlowsmallcaps{#1}}}
\clearpage
\begingroup
    \let\clearpage\relax
    \let\cleardoublepage\relax
    \pdfbookmark[1]{\listfigurename}{lof}
    \listoffigures

    \vspace{8ex}

    \pdfbookmark[1]{\listtablename}{lot}
    \listoftables

    \vspace{8ex}

    \pdfbookmark[1]{\lstlistlistingname}{lol}

    \vspace{8ex}


\endgroup

%% file: Chapters/Chapter01.tex
\chapter{Introduction}\label{ch:introduction}

The human brain is one of the most complex systems in the universe.
Its approximately 86 billion neurons \citep{azevedo2009equal} and 100--500 trillion synapses \citep{drachman2005we} are capable of abstract reasoning, pattern recognition, memorization, and sensory experience---while consuming only about \SI{20}{\watt} of power \citep{sokoloff1960metabolism,drubach2000brain}.

An early goal of artificial intelligence has been to construct systems that exhibit similar intelligent traits \citep{turing1948intelligent}.
One of the proposed methods was to emulate the network of biological neurons in the human brain using simple units called perceptrons \citep{rosenblatt1958perceptron}.
These perceptrons were inspired by Hebbian learning \citep{hebb1949organization}, which was a new (and later corroborated) theory on biological learning.
After the popularization in the 1980s of trainable Hopfield networks \citep{hopfield1982neural} and backpropagation \citep{rumelhart1986learning}, which enabled learning linearly inseparable tasks, artificial neural networks (ANNs) and the connectionist approach were embraced with a new appreciation.

These ANNs are networks of small computational units that can be trained to perform specific pattern recognition tasks.
Backpropagation has proven to work well in training ANNs with multiple layers, most popularly in the field of deep learning (DL), which has become a dominant field in artificial intelligence.
This popularity is partly due to exponentially increasing computing power and data storage capabilities, as well as the rise of the Internet, which has also provided ample training data.
Some variations on ANNs have shown to improve learning performance, such as using convolutional (CNN) and recurrent (RNN) neural networks, both of which are, like the perceptron, inspired by the architecture of the human brain \citep{hubel1968receptive, fukushima1982neocognitron,lecun1995convolutional,lukovsevivcius2009reservoir}.
These types of networks approach or exceed human level performance in some areas \citep{schmidhuber2015deep}.

\paragraph{Energy limits}
	However, DL-based methods are starting to show diminishing returns; training some state-of-the-art models can require so much data and computing power that only a small number of organizations has the resources to train and deploy them.
	For example, one of the current top submissions of the ``Labeled Faces in the Wild'' face verification task is a deep ANN by Paravision that was trained using a dataset of 10 million face images of 100 thousand individuals\footnote{See \texttt{http://vis-www.cs.umass.edu/lfw/results.html\#everai}. Last accessed January 2021.}.
	Beside very large datasets, deep ANNs also require a significant amount of power to train.
	For instance, ResNet \citep{he2016deep} has been trained for 3 weeks on a 8-GPU server consuming about 1 GWh.
	This high power consumption precludes computations in mobile low-power or small-scale devices, which now require at least a connection to a cloud computing server.

	The energy consumption of DL contrasts strongly with that of the human brain, which can learn patterns using far less energy and data.
	This is because despite the biologically inspired foundation, deep ANNs are fundamentally different from the brain, which is an inherently time-dependent dynamical system \citep{sacramento2018dendritic, wozniak2020deep} that relies on biophysical processes, recurrence, and feedback of its physical substrate for computation \citep{sterling2015principles,bhalla2014molecular}.
	Deep ANNs are implemented on von Neumann architectures \citep{von1993first}, \ie, a system with a central processing unit (CPU) and separate memory, which are significantly different from the working model of the brain \citep{schuman2017survey}.

	One reason for the inefficiency of deep ANNs is that their implementations suffer from the von Neumann bottleneck \citep{zenke2021brain}, which involves a limited throughput between the CPU and memory---a data operation cannot physically co-occur with fetching instructions to process that data because they share the same communication system.
	Parallelization on GPUs has alleviated this bottleneck to some extent, but the human brain is more efficient as it is embedded in a physical substrate whose neurons operate and communicate fully in parallel \citep{a2017parallel} using sparsely occurring \emph{spikes} \citep{bear2020neuroscience}, and where no explicit data processing instructions exist.
	A spike can be represented as a binary value which causes the synapse to increase the membrane potential in the efferent neuron to change by a fixed value \citep{bear2020neuroscience}.
	Connections in ANNs are represented abstractly by large weight matrices, which are all multiplied with neuron activation values at every propagation cycle.
	In the brain, a synapse spikes sparsely and thereby saves energy while conveniently including an informative temporal component.

	A second reason is that backpropagation requires two passes over the ANN: the first to compute the network output given an input, and the second to propagate the output error back into the network to move the weights between neurons in the direction of the negative gradient.
	Backpropagation in RNNs is often performed by unrolling the network in a feedforward ANN in a process called backpropagation through time (BPTT).
	The human brain, in contrast, is unlikely to use backpropagation, BPTT, or gradients of the output error \citep{lillicrap2019backpropagation}.

\paragraph{Spiking neural networks}
	Spiking neural networks (SNNs) \citep{maass1997networks,gerstner2002spiking} are another step towards biological plausibility of connectionist models.
	SNNs use neurons that do not relay continuous activation values at every propagation cycle, but spike once when they reach a threshold value.
	This makes SNNs potentially much more energy efficient than ANNs that use backpropagation.
	SNNs are competitive to ANNs in terms of accuracy and computational power, as well in their ability to display precise spike timings \citep{lobo2020spiking}.
	Their sparse firing regimes also offer improved interpretability of their behavior as compared to traditional ANNs \citep{soltic2010knowledge}, which is desired in areas such as medicine or aviation.

	However, SNNs have not been as popular as ANNs.
	One reason for this is that spike-based activation is not differentiable.
	As a consequence, backpropagation cannot be directly used to move in the negative direction of the error gradient, although some attempts have been made to bridge this divide \citep{bohte2002error,hong2010cooperative,xu2013supervised,ourdighi2016efficient,lee2016training,sacramento2018dendritic,bellec2019biologically,whittington2019theories} and to make backpropagation more biologically plausible.

	Similarly, it has been demonstrated that approximations of BPTT can be applied to recurrent SNNs (RSNNs) \citep{huh2017gradient,bellec2018long}.
	Both single- and multi-layer SNNs have shown good performance in visual processing \citep{escobar2009action,kheradpisheh2018stdp,liu2017fast} and speech recognition \citep{tavanaei2017bio,dong2018unsupervised}.
	However, none of these algorithms are biologically plausible.
	While DL was rapidly becoming popular during the 2010s, there was no clear learning algorithm for SNNs that could compete with ANNs.
	A second reason for the relative unpopularity of SNNs is that they are generally emulated in von Neumann architectures, undermining their energy efficiency advantages.

\paragraph{Neuromorphic computing}  
	SNN learning algorithms are particularly useful in the upcoming field of neuromorphic computing (NC) \citep{mitra2008real}, in which analog very-large-scale integration (VLSI) systems are used to implement neural systems.
	On the surface, it can be understood as running neural networks not abstracted in a digital system, but physically embedded in a dedicated analog medium.
	A central advantage of NC is energy efficiency \citep{hasler1990vlsi,lee1990parallel,tarassenko1990real}.
	This energy efficiency, combined with NC's massive parallelism \citep{monroe2014neuromorphic}, makes VLSIs particularly relevant for implementing SNNs.

	Like SNNs, neuromorphic systems typically use sparse, event-based communication between devices
	and physically colocalized memory and computation \citep{sterling2015principles,neftci2018data}.
	Although colocalized memory and computation has also been implemented in digital machines, such as Google's TPU\footnote{See \texttt{https://cloud.google.com/tpu/docs/tpus}. Last accessed January 2021}, Graphcore's IPU\footnote{See \texttt{https://www.graphcore.ai/products/ipu}. Last accessed January 2021}, or Cerebras' CS-1\footnote{See \texttt{https://cerebras.net/product/\#chip}. Last accessed January 2021}, neuromorphic systems are more efficient for running ANNs \citep{merolla2014million,rajendran2019low}.
	The energy consumption of CMOS artificial neurons is several orders of magnitude lower than that of neurons in an ANN, and even 2--3 times lower than the energy consumption of biological neurons \citep{elbez2020progressive}, offering a possible escape from the increasing energy costs of DL models.

	Because of this massive parallelism, high energy efficiency, and good ability to implement cognitive functions, neuromorphic systems are attracting strong interest.
	In particular, SNNs emerged as an ideal biologically inspired NC paradigm for realizing energy-efficient on-chip intelligence hardware \citep{merolla2014million,davies2018loihi}, suitable for running fast and complex SNNs on low-power devices.
	For instance, a competitive image classification performance was reached with a 6-order of magnitude speedup in a leaky integrate-and-fire (LIF) SNN in field-programmable gate arrays, compared to digital simulations \citep{zhang2020low}.

\paragraph{Biological learning}
	To run an SNN on neuromorphic hardware, a \emph{local} and \emph{online} learning algorithm is needed.
	The precondition of locality refers to the idea that a neuron or synapse can only access information or communication with which it is physically connected.
	For instance, the inner state of a neuron can only be influenced by itself, or by the spikes it receives from afferent neurons.
	Similarly, a synapse can only spike or change its weight based on signals from the afferent and efferent neuron.
	This is a direct consequence of the colocalization of processing and memory.
	The precondition of being online can be regarded as temporal locality---neurons and synapses can only access information that physically exists at the same point in time.
	They cannot access information about past or future events, except if explicit memory traces of a past event are retained over time.
	In that case, past events can affect the neuron's current behavior.

	The brain also adheres to these two constraints.
	Some of the more common learning rules in ANNs are based on a form of Hebbian learning, which is a major factor in biological learning and memory consolidation \citep{schuman2017survey}.
	Classical Hebbian learning is often summarized by ``cells that fire together, wire together'', if there is a causal relationship between these cells, such as a postsynaptic potential on a connecting synapse.
	Direct application of Hebbian learning in a spiking neural network will generally lead to a positive feedback loop, because ``wiring cells together'', or increasing the synaptic strength, will in turn increase the likelihood that they also fire together \citep{zenke2017temporal}.
	Furthermore, classical Hebbian learning describes no way for a synapse to weaken.

	Spike-timing-dependent plasticity (STDP) \citep{abbott2000synaptic,caporale2008spike} is a type of Hebbian learning that incorporates temporal causality on a synapse from neuron $A$ to neuron $B$: if $B$ spikes right after neuron $A$, then the synapse is strengthened, but if $B$ spikes right before $A$, it is weakened.
	It is widely known that STDP is a fundamental learning principle in the human brain \citep{kandel2000principles,caporale2008spike}, including perceptual systems in the sensory cortex \citep{huang2014associative}.
	STDP by itself can be used as an unsupervised learning algorithm or to forming associations in classical conditioning \citep{diehl2015unsupervised,kim2018demonstration}.
	Furthermore, it has been demonstrated to form associations between memory traces in SNNs, which are crucial for cognitive brain function \citep{pokorny2020stdp}.
	To allow supervised learning, or operant conditioning, a learning signal is required to influence the direction of the synapse weight change: a positive learning signal will reinforce the association (long-term potentiation), and a negative learning signal weakens it (long term depression) \citep{lobov2020spatial}.
	STDP with a learning signal is known as reward-modulated STDP (R-STDP) \citep{legenstein2008learning} in the field of SNNs and three-factor Hebbian learning in neuroscience \citep{fremaux2016neuromodulated}.
	Three-factor Hebbian learning has been demonstrated to outperform its classical two-factor counterpart in a localization-and-retrieval task \citep{porr2007learning}.
	A possible reason for this performance difference is that modulatory signals ``may provide the attentional and motivational significance for long-term storage of a memory in the brain'' and stabilize classical Hebbian learning \citep{bailey2000heterosynaptic}.

	Neurotransmitters are used to modulate the learning signal in the brain.
	Dopamine, for instance, which has a central behavioral and functional role in the primary motor cortex \citep{barnes2005activity,dang2006disrupted}, has been shown to modulate synapses through dendritic spine enlargement during a very narrow time window \citep{dang2006disrupted}.
	It is behaviorally related to novelty and reward prediction \citep{li2003dopamine,schultz2007behavioral} by gating neuroplasticity of corticostriatal \citep{reynolds2001cellular,reynolds2002dopamine} and ventral tegmental (VTA) synapses \citep{bao2001cortical}.
	In the VTA, dopaminergic neurons respond to learning signals in a highly localized manner that is specific for local populations of neurons \citep{engelhard2019specialized}.
	This is also the case in other areas of the midbrain \citep{roeper2013dissecting}.

	However, R-STDP by itself does not solve the credit assignment problem, which relates to neuromodulation of synapses after a learning signal is presented with some delay.
	In that case, when the learning signal is presented, the neurons have long spiked, and it is not clear which synapses elicited the behavior that is rewarded or punished.
	Recent research suggests that the brain uses \emph{eligibility traces} \citep{izhikevich2007solving,florian2007reinforcement} to solve the credit assignment problem \citep{stolyarova2018solving,gerstner2018eligibility}.
	In particular, the synaptic CaMKII protein complex is activated during the induction of long-term potentiation (LTP) of biological synapses if the presynaptic neuron spikes shortly before a postsynaptic neuron \citep{sanhueza2013camkii}.
	This LTP is maintained over behavioral time spans, and gradually fades.
	When followed by a learning signal in the form of a neurotransmitter, synaptic plasticity is induced \citep{yagishita2014critical,cassenaer2012conditional,gerstner2018eligibility}.

	Over the past decade, eligibility traces have been researched in the context of a wide range of topics, such as biological learning, spiking neural networks, and neuromorphic computing.
	Synaptic plasticity was demonstrated using eligibility traces in deep feedforward SNNs \citep{zenke2018superspike,neftci2017event,kaiser2020synaptic} and could be implemented in feedforward VLSIs.
	In \citet{zenke2018superspike} it is asserted that these methods are also applicable for RSNNs.
	Eligibility traces have also been shown to solve difficult credit assignment problems in SNNs using R-STDP \citep{legenstein2008learning, bellec2020solution} and in RNNs \citep{he2015distinct}, and have a predictable learning effect \citep{legenstein2008learning}.

\paragraph{Eligibility propagation}
	Eligibility propagation (e-prop) \citep{bellec2020solution} is a local and online learning algorithm for RSNNs that can be mathematically derived as an approximation to BPTT (see also Section \ref{sec:derivefromBPTT}).
	The main aspect that distinguishes e-prop from other eligibility trace--based algorithms is that the particular computation of the eligibility trace depends on multiple hidden states of a neuron.
	The property that a neuron can have multiple hidden states means that there are many types of neuron models that can be used in e-prop.

	In e-prop, the learning signal is a local variation on random broadcast alignment, which propagates the error directly back onto the neurons with a random weight, resembling the function of a neuromodulator in the brain.
	This has been suggested to provide a diversity of feature detectors for task-relevant network inputs \citep{bellec2020solution}.
	This form of broadcast alignment can perform as effectively as backpropagation in some tasks in feedforward ANNs \citep{lillicrap2016random,nokland2016direct} and multi-layer SNNs \citep{samadi2017deep,clopath2010connectivity}, but performs poorly in deep feedforward ANNs for complex image classification tasks \citep{bartunov2018assessing}.

	The local and online properties of e-prop make it a biologically plausible learning algorithm that can be implemented on VLSIs.
	E-prop has been demonstrated to work for a large variety of tasks, including classifying phones (\ie, speech sounds), for which it performs competitively with RNNs that use BPTT and the popular LSTM neuron model \citep{graves2013speech}.

	The fading eligibility trace in e-prop is similar to STDP in that the weight change is smaller if there is a longer delay between a presynaptic and postsynaptic spike.
	However, e-prop is essentially independent of STDP, because it does not explicitly relate the order of the pre- and postsynaptic spike to the synaptic weight update.
	However, in \citet{bellec2020solution} e-prop is remarked to start showing STDP-like properties if the synaptic delay of a spike is prolonged.

	So far, only the LIF and adaptive LIF (ALIF) neuron models have been used in e-prop, which do not show STDP-like properties by default.
	In \citet{traub2020learning}, a functional modification was made to the LIF model such that STDP can occur.
	In particular, STDP occurs when the neuron model provides a negated gradient signal in the case when a presynaptic signal arrives too late.
	This resembles the biological phenomenon of error-related negativity (ERN) \citep{nieuwenhuis2001error}, which is a negative brain response that immediately follows an erroneous behavioral response and peaks after 80--150 ms with an amplitude that depends on the intent and motivation of a person.
	\citet{traub2020learning} also showed this effect for the Izhikevich neuron \citep{izhikevich2003simple}.
	However, these STDP-modified neurons were shown only in a single-synapse demo to illustrate the STDP properties, not in a full learning task.

\paragraph{Multi-layer RSNNs}

	The discovery of backpropagation allowed gradient descent--based training of multi-layer ANNs, which significantly increased their performance.
	Although it is unlikely to use backpropagation, the human brain is hierarchically structured such that early layers process simple information and deep layers process more abstract information.
	Similarly, multi-layer CNNs also show higher levels of abstraction in deeper layers of the network.
	For instance, early convolutional filters identify lines and edges, while deeper filters identify more complex shapes.
	In RNNs, stacking recurrent layers results in a similar abstraction---but it is temporal instead of spatial \citep{hermans2013training,gallicchio2017deep}.
	Deeper RNN layers exhibit slower time dynamics and longer memory spans than shallow layers \citep{gallicchio2018short}, suggesting that they ignore small variations in the input signal and integrate larger temporal patterns.
	It is unclear if these findings extrapolate to RSNNs.

\paragraph{Research objectives}
	State-of-the-art SNN learning algorithms perform well on a variety of tasks, but have so far not shown the efficiency and learning performance of the human brain.
	SNNs are most efficient when embedded in a neuromorphic system, requiring a learning algorithm that is local and online.
	E-prop is an example of such an algorithm, but it has not yet been used in conjunction with neuron models other than LIF and ALIF.
	These neuron models do not show STDP-like behavior, which is a fundamental learning principle in the brain, and has a close connection to biological eligibility traces, and may therefore improve the accuracy and efficiency of e-prop.
	For this reason, in this research I continue the trend of emulating biological processes by for the first time modifying the e-prop network to use neuron models that show STDP-like behavior.
	Analyzing the effects of including STDP-like behavior to the neuron models in an e-prop network is the primary research objective in this report.

	Two neuron models that display STDP are used.
	The first model is the ALIF-STDP, which is a new crossover neuron model of the ALIF neuron (used in \citet{bellec2020solution}) and the STDP-LIF neuron (derived but not verified for e-prop in \citet{traub2020learning}).
	The second STDP-like neuron model is the Izhikevich neuron model, which was also derived in \citet{traub2020learning}, and is slightly modified in this research to produce stable eligibility traces over time.

	So far, only the performance of e-prop models with a single fully-connected pool of neurons has been described.
	Whereas multi-layered CNNs and RNNs can sometimes process abstract information more easily, it is not clear if this also holds for SNNs or e-prop models.
	The secondary research objective is analyzing the effects of a multi-layered e-prop architecture.

\paragraph{Structure of this report}
	In the remainder of this report, the e-prop framework is described in Chapter \ref{ch:relatedwork}.
	Then, Chapter \ref{ch:method} describes the method used to implement the TIMIT phone classification task and modify the e-prop algorithm to a multi-layer framework with different neuron models, particularly the STDP-ALIF and Izhikevich models.
	This modified e-prop framework is implemented and experimentally verified.
	The results are presented and discussed in Chapter \ref{ch:discussion}.
	These results show that including STDP in ALIF neuron models can indeed improve the learning performance and leads to a higher classification accuracy.
	However, this does not hold for the Izhikevich neuron, suggesting that this neuron model is not suited for e-prop in its current form.
	Furthermore, the use of multiple stacked recurrent layers slows down the learning speed, and so does not provide an efficient e-prop architecture.
	Finally, Chapter \ref{ch:conclusion} summarizes and concludes this report.

%% file: Chapters/Chapter02.tex
\chapter{Theoretical Framework}\label{ch:relatedwork}

    This chapter describes the theoretical framework of eligibility propagation expounded in previous literature, which is then developed further in Chapter \ref{ch:method}.

    \section{Eligibility propagation model}
        In \citet{bellec2020solution}, an eligibility propagation (e-prop) model $\mathcal{M}$ of a neuron $j$ in a feedforward or recurrent network is defined by a tuple $\left<M, f\right>$,
        where $M$ is a function
        \begin{equation}\label{eq:model}
        \mathbf{h}^t_j = M\left(\mathbf{h}_j^{t-1}, \mathbf{z}^{t-1}, \mathbf{x}^t, \mathbf{W}_j\right)
        \end{equation}
        that defines the hidden state $\mathbf{h}_j^t$ at a discrete time step $t$, where $\mathbf{z}^{t-1}$ is the observable state of all neurons at the previous time step (\ie, the binary spike values), $\mathbf{x}^t$ is the model input vector at time $t$, and $\mathbf{W}_j$ is the weight vector of afferent (\ie, ``incoming'') synapses.
        The hidden state of a neuron contains all variables that are used for a specific neuron type, \eg, an activation value, or a variable that models a neuron's refractory period after a spike.
        In short, Equation \ref{eq:model} indicates that the hidden state is affected primarily by spikes of other neurons $\mathbf{z}^{t-1}$, and the current input to the model $\mathbf{x}^t$, which are both weighted by trainable network weights $\mathbf{W}^\text{rec}_j \subset \mathbf{W}_j$ and $\mathbf{W}^\text{in}_j \subset \mathbf{W}_j$, respectively.

        The function $f$ in $\mathcal{M}$ describes the update of the observable state of a neuron $j$ at time $t$:
        \begin{equation}\label{eq:observable}
        z^t_j = f\left(\mathbf{h}_j^t\right),
        \end{equation}
        such that the spikes elicited by a neuron only depend on its hidden state.
        For instance, a neuron $j$ may spike at time $t$ (\ie, $z^t_j = 1$) if its activity, which is contained in the hidden state, reaches a threshold value.

        The purpose of an e-prop model is that it can be trained to perform a learning task, such as classification.
        As described in the remainder of this chapter, the weight matrix $\mathbf{W}$, which comprises the weight values of all synapses in the model, is trained such that the input vectors $\mathbf{x}^t$ yield a good prediction of the classes they belongs to.

        The formalizations described in Equations \ref{eq:model} and \ref{eq:observable} indicate that e-prop is a \emph{local} training method, because a neuron's observable state depends only on its own hidden state, which in turn depends only on observable signals that are directly connected to it.
        E-prop is also an \emph{online} training method, because both the hidden and observable state of a neuron depend only on information that is still available; the observable state is updated at the same time step as the hidden state, and the hidden state is updated according to information which is then present in the afferent neurons.

        The precise formulations of $M$ and $f$ depend on the neuron models that are used in the e-prop model.

    \section{Neuron models}\label{sec:alif}

        \paragraph{LIF neuron}
            In \citet{bellec2020solution}, the LIF neuron model is formulated in the context of e-prop, along with a variant (viz. ALIF) that has an adaptive threshold based on the neuron's spiking frequency.
            The observable state of a LIF model is given by
            \begin{equation}\label{eq:heavisideLIF}
            z^t_j = H\left(v_j^t-v_\text{th}\right),
            \end{equation}
            where $H$ is the Heaviside step function, $v^t_j$ is the activity of neuron $j$ at discrete time $t$, and $v_\text{th}$ is the threshold constant.
            (Note that this and all other used hyperparameters are listed in Table \ref{tab:hparams}.)
            From Equation \ref{eq:heavisideLIF} it follows that a neuron spikes ($z^t_j = 1$) if its activity reaches the activity threshold, and remains silent ($z^t_j = 0$) otherwise.
            These spikes are the only communication between neurons in the e-prop model.

            The hidden state $h^t_j$ of a LIF neuron model contains only an activity value $v^t_j$ that evolves over time according to the equation
            \begin{equation}\label{eq:alifV}
            v^{t+1}_j = \alpha v_j^t + \sum_{i\neq j}W^\text{rec}_{ji}z_i^t + \sum_i W^\text{in}_{ji}x_i^{t+1} - z_j^tv_
            \text{th},
            \end{equation}
            where $W^\text{rec}_{ji}$ is a synapse weight from neuron $i$ to neuron $j$, and $\alpha$ is a constant decay factor.
            In Equation \ref{eq:alifV}, the first term models the decay of the activity value over time.
            The second and third term model the input of the neuron from other neurons, or from the input to the network, respectively.
            The fourth term ($-z^t_jv_\text{th}$) ensures that the activity of the neuron drops when it spikes.
            Furthermore, $z^t_j$ is explicitly fixed to 0 for $T^\text{refr}$ time steps after a spike to model neuronal refractoriness.

            In biological neurons, the refractory period consists of an ``absolute'' phase, during which eliciting a new spike is impossible, and a subsequent ``relative'' phase, during which the threshold is temporarily increased \citep{purves2008neuroscience}.
            Clamping $z^t_j$ to 0 emulates only this absolute phase, and is therefore only a crude approximation to model biological refractoriness.
            The refractory period is built into the equations of the Izhikevich neuron model described in Section \ref{sec:izhikevich}, which is therefore arguably a more biologically plausible neuron model.

        \paragraph{ALIF neuron}
            The ALIF neuron model introduces a threshold adaptation variable $a^t_j$ to the hidden state of the LIF neuron, such that $\mathbf{h}^t_j \eqdef \left[v^t_j, a^t_j\right]$.
            In an ALIF neuron, the spiking threshold increases after a spike, and otherwise decreases back to a baseline threshold $v_\text{th}$ in the continued absence of spikes.

            This resembles \emph{spike frequency adaptation} (SFA), a common feature of neocortical pyramidal neurons \citep{benda2003universal}.
            SFA is a homeostatic control mechanism that affects the spiking frequency based on the recent spiking activity, such that neurons that spike relatively infrequently become more sensitive, and vice versa.
            \citet{ahmed1998estimates} found that a single time constant is a good fit to characterize the threshold's exponential decay to a steady state.

\begin{figure}[!ht]
                \centering
                \includegraphics[width=\linewidth]{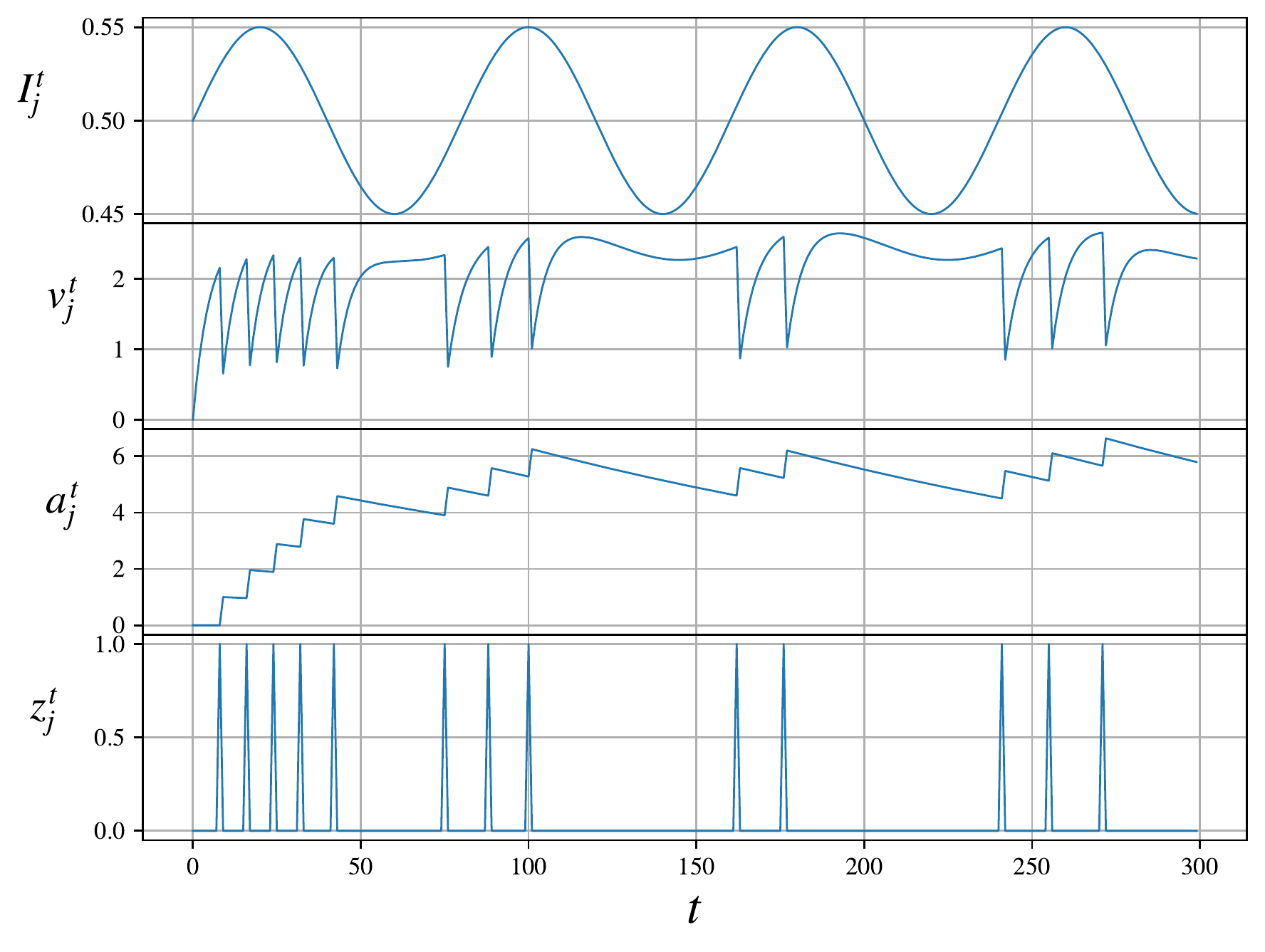}
                \caption[ALIF neuron simulation]{A simulated ALIF neuron $j$ receives a sinusoidal input $I$ for 300 time steps $t$. This figure illustrates the adaptive threshold $a$, which increases at every spike $z$, requiring a higher activity $v$ for a next spike. When a spike occurs, $v$ decreases by $v_\text{th}$. Note that the first wave of the sinusoid elicits a stronger spike train than subsequent waves, demonstrating the homeostatic effect of the adaptive threshold. Note also that on a short time scale, spikes tend to occur primarily in the upward phases of the sinusoid, suggesting that ALIF neurons are well-suited to respond to changes in the input signals.}
                \label{fig:simplealif}
            \end{figure}

            The observable state of an ALIF neuron is therefore described by
            \begin{equation}\label{eq:alifZ}
            z^t_j = H\left(v_j^t - v_\text{th} - \beta a^t_j\right)
            \end{equation}
            and
            \begin{equation}\label{eq:alifA}
            a^{t+1}_j = \rho a^t_j + z^t_j,
            \end{equation}
            where $\rho < 1$ is an adaptation decay constant and $\beta \leq 0$ is an adaptation strength constant.
            Equation \ref{eq:alifA} indicates that the adaptive threshold increases at every spike, and decays back to $v_\text{th}$ in the absence of spikes.
            The decay factor $\rho$ of the threshold adaptation is higher than the decay factor $\alpha$ of the neuron activity, such that the immediate firing behavior of a neuron is affected on a shorter time scale than the threshold adaptation, which is better suited to reflect the working memory of a neuron and track longer temporal dependencies in the input data than the activation decay.
            The interaction between the neuron activity, adaptive threshold and spiking behavior is illustrated in Figure \ref{fig:simplealif}.

            The LIF neuron is a spacial case of an ALIF neuron, for which $\beta=0$, effectively canceling the effect of the threshold adaptation value $a^t_j$ on the observable state $z^t_j$ in Equation \ref{eq:alifZ}.
            Therefore, only the e-prop derivations for the ALIF neurons will be described in the following sections.
            From this point, references to LIF neurons in this report will refer to this special case.

    \section{Network topology}
        The e-prop network structure as used in this report consists of the following main components:
        \begin{enumerate}
            \item An input layer $\mathbf{x}^t$.
            \item A recurrent layer containing $N$ neurons that are connected to all other neurons in this layer by weights $\mathbf{W}^\text{rec}$. This layer is also connected to the input layer by weights $\mathbf{W}^\text{in}$.
            \item An output layer $\mathbf{y}^t$ connected to the recurrent layer by weights $\mathbf{W}^\text{out}$.
        \end{enumerate}

        Since one of the goals of this report is to evaluate multi-layer topologies, the recurrent layer component is modified in Section \ref{sec:ml_arch} to support architectures with a feedforward series of recurrent layers.

\begin{figure}[!ht]
            \myfloatalign
            \includegraphics[trim=0 25cm 0 0, clip, width=\linewidth]{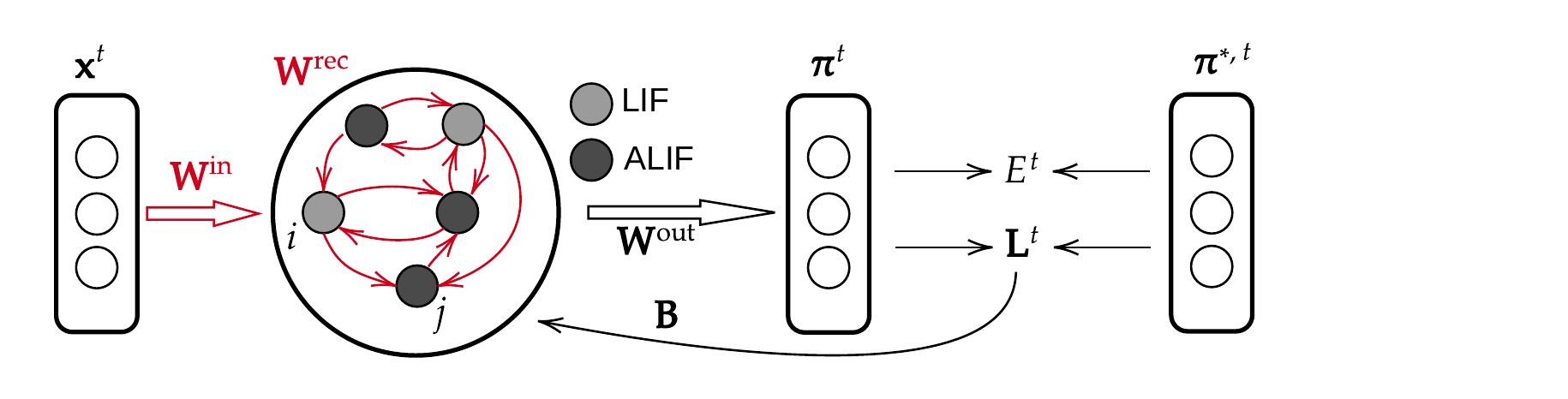}\caption[Single-layer architecture illustration]{A basic illustration of a single-layer network architecture.
            An input vector $\mathbf{x}^t$ at time step $t$ is fed to a pool of $N$ recurrent neurons with hidden states $\mathbf{h}^t$ and observable states $\mathbf{z}^t$ through input weights $\mathbf{W}^\text{in}$. The recurrent weights $\mathbf{W}^\text{rec}$ connect neurons with each other, but no self-loops exist. A randomly selected 25\% of these neurons is a LIF neuron (\ie, $\beta = 0$) and the others are ALIF neurons. The output weights $\mathbf{W}^\text{out}$ process the observable states of the neurons through a softmax function into a logits layer $\mathbf{\pi}^t$. These logits are compared with the target output $\mathbf{\pi}^{*, t}$ and multiplied with broadcast weights $\mathbf{B}^t$ to obtain a learning signal $L_j^t$ for every neuron $j$ in the pool. Note that weights illustrated in red are e-prop weights, \ie, they track eligibility traces.}
            \label{fig:topology-sl}
          \end{figure}

        An input vector $\mathbf{x}^t$ at time step $t$ is fed to a pool of $N$ recurrent neurons with hidden states $\mathbf{h}^t$ and observable states $\mathbf{z}^t$ through input weights $\mathbf{W}^\text{in}$.
        The recurrent weights $\mathbf{W}^\text{rec}$ connect neurons with each other, but no self-loops exist.
        Therefore, the recurrent neurons also receive inputs from the observable states of the afferent neurons.
        25\% of these neurons are LIF neurons (\ie, $\beta = 0$) and the others are ALIF neurons.
        The output weights $\mathbf{W}^\text{out}$ process the observable states of the neurons through a softmax function into a logits layer $\mathbf{\pi}^t$.
        These logits are compared with the target output $\mathbf{\pi}^{*, t}$ and multiplied with broadcast weights $\mathbf{B}^t$ to obtain a learning signal $L_j^t$ for every neuron $j$ in the pool.
        Figure \ref{fig:topology-sl} illustrates the basic architecture of a single-layer e-prop model.

        Like in \citet{bellec2020solution}, weights are initialized by sampling them from a Gaussian distribution $\mathcal{N}\left(0, \sqrt{N}\right)$ where $N$ is the number of afferent neurons.
        For instance, the weights $\mathbf{W}^\text{in}$ between the input and the first layer are sampled from $\mathcal{N}\left(0, \sqrt{39}\right)$ if there are 39 input features.
        Likewise, each of the neurons has $N-1$ afferent recurrent weights, so the recurrent weights within a layer are sampled from $\mathcal{N}\left(0, \sqrt{N-1}\right)$.

        A randomly selected 80\% of synaptic weights is then set to a value of 0, as well as the synapses that connect a neuron to itself, rendering them ineffective.

    \section{Deriving e-prop from RNNs}\label{sec:derivefromBPTT}
        Eligibility propagation is a local and online training method that can be derived from backpropagation through time (BPTT).
        In BPTT, an RNN is unfolded in time, such that the backpropagation method used in feedforward neural networks can be applied to compute the gradients of the cost with respect to the network weights.

        In this section, the main equation of e-prop
        \begin{equation}
        \frac{dE}{dW_{ji}} =
        \sum_t\frac{dE}{dz_j^t}\cdot\left[\frac{dz_j^t}{dW_{ji}}\right]_\text{local},
        \end{equation}
        where $\cdot$ denotes the dot product, is derived from the classical factorization of the loss gradients in an unfolded RNN as in \citet{bellec2020solution}:
        \begin{equation}\label{eq:clafac}
        \frac{dE}{dW_{ji}} = \sum_{t' \leq T}\frac{dE}{d\mathbf{h}_j^{t'}}\cdot\frac{\partial \mathbf{h}_j^{t'}}{\partial W_{ji}},
        \end{equation}
        where summation indicates that weights are shared.
        Recall that for ALIF neurons, $\mathbf{h}^t_j \eqdef \left[v^t_j, a^t_j\right]$.
        This is also the true for ALIF neurons that use $\beta=0$ to disable their threshold adaptability.

        By applying the chain rule, the first factor $\frac{dE}{d\mathbf{h}_j^{t'}}$ can be decomposed into a series of learning signals $L_j^t = \frac{dE}{dz_j^t}$ and local factors $\frac{\partial\mathbf{h}_j^{t-t'}}{\partial\mathbf{h}_j^t}$ for all $t$ starting from the event horizon $t'$, which is the oldest time step that information is used from:

        \begin{equation}\label{eq:rec}
        \frac{dE}{d\mathbf{h}_j^{t'}} = \underbrace{\frac{dE}{dz_j^{t'}}}_{L^{t'}_j} \frac{\partial z_j^{t'}}{\partial\mathbf{h}_j^{t'}} + \frac{dE}{d\mathbf{h}_j^{t'+1}}\frac{\partial\mathbf{h}_j^{t'+1}}{\partial\mathbf{h}_j^{t'}}.
        \end{equation}
        Note that this equation is recursive.
        If Equation \ref{eq:rec} is substituted into the classical factorization (Equation \ref{eq:clafac}), the full history of the synapse $i\rightarrow j$ is integrated, and a recursive expansion is obtained that has $\frac{dE}{d\mathbf{h}^{T+1}_j}$ as its terminating case:
        \begin{align}
        \frac{dE}{dW_{ji}} &= \sum_{t'}\left(L_j^{t'}\frac{\partial z_j^{t'}}{\partial\mathbf{h}_j^{t'}} + \frac{dE}{d\mathbf{h}_j^{t'+1}}\frac{\partial\mathbf{h}_j^{t'+1}}{\partial\mathbf{h}_j^{t'}}\right)\cdot\frac{\partial\mathbf{h}_j^{t'}}{\partial W_{ji}}\\
        &= \sum_{t'}\left(L_j^{t'}\frac{\partial z_j^{t'}}{\partial\mathbf{h}_j^{t'}} + \left( L^{t'+1}_j \frac{\partial z_j^{t'+1}}{\partial\mathbf{h}_j^{t'+1}} + (\cdots)\frac{\partial\mathbf{h}_j^{t'+2}}{\partial\mathbf{h}_j^{t'+1}}  \right) \frac{\partial\mathbf{h}_j^{t'+1}}{\partial\mathbf{h}_j^{t'}}\right)\cdot\frac{\partial\mathbf{h}_j^{t'}}{\partial W_{ji}}.
        \end{align}

        The recursive parenthesized factor can be written into a second factor indexed by $t$:
        \begin{equation}
        \frac{dE}{dW_{ji}} = \sum_{t'}\sum_{t\geq t'}L^t_j\frac{\partial z_j^t}{\partial\mathbf{h}_j^t}\frac{\partial\mathbf{h}^t_j}{\partial\mathbf{h}_j^{t-1}} \cdots \frac{\partial\mathbf{h}_j^{t+1}}{\partial\mathbf{h}_j^{t'}}\cdot\frac{\partial\mathbf{h}_j^{t'}}{\partial W_{ji}}.
        \end{equation}

        By exchanging the summation indices, the learning signal $L_j^t$ is pulled out from the inner summation.

        Within the inner summation, the terms $\frac{\partial\mathbf{h}_j^{t+1}}{\partial\mathbf{h}_j^t}$ are collected in an \emph{eligibility vector} $\bm{\epsilon}^t_{ji}$ and multiplied with the learning signal $L^t_j$ at every time step $t$.
        This is crucial for understanding why e-prop is an online training method---local gradients are computed based on traces that are directly accessible at the current time step $t$, and the eligibility vector operates as a recursively updated ``memory'' to track previous local hidden state derivatives:

        \begin{equation}
        \bm{\epsilon}^t_{ji} = \frac{\partial\mathbf{h}_j^{t}}{\partial\mathbf{h}_j^{t-1}}\cdot\bm{\epsilon}^{t-1}_{ji} + \frac{\partial\mathbf{h}^t_j}{\partial W_{ji}}.
        \end{equation}

        This is why the $\rho$ and $\alpha$ parameters, which define the decay rate in hidden states and the corresponding eligibility vectors, should be set according to the required working memory in the learning task.
        The eligibility vector and the hidden state have the same dimension: $\left\{\bm{\epsilon}^t_{ji}, \mathbf{h}^t_j\right\} \subset \mathbb{R}^d$, where $d=2$ for the ALIF and Izhikevich neuron models.

        The \emph{eligibility trace} $e^t_{ji}$ is a product of $\frac{\partial z_j^t}{\partial \mathbf{h}_j^t}$ and the eligibility vector, resulting in the gradient that can be immediately applied at every time step $t$, or accumulated and integrated locally on a synapse (see Section \ref{sec:eprop_grd} for details):
        \begin{equation}\label{eq:main_eprop}
        \frac{dE}{dW_{ji}} = \sum_t\frac{dE}{dz_j^t}\underbrace{\frac{\partial z_j^t}{\partial\mathbf{h}_j^t}\underbrace{\sum_{t\geq t'}\frac{\partial\mathbf{h}^t_j}{\partial\mathbf{h}_j^{t-1}} \cdots \frac{\partial\mathbf{h}_j^{t+1}}{\partial\mathbf{h}_j^{t'}}\cdot\frac{\partial\mathbf{h}_j^{t'}}{\partial W_{ji}}}_{\bm{\epsilon}_{ji}^t}}_{e^t_{ji}}.
        \end{equation}
        This is the main e-prop equation.

    \section{Learning procedure}

        The e-prop equation (Equation \ref{eq:main_eprop}) can be applied to any neuron type with any number of hidden states.
        In this section, the derivation for ALIF neurons will be detailed.

        \subsection{Eligibility trace}
            Recall the hidden state update equations from Section \ref{sec:alif}:
            \begin{equation*}
            v^{t+1}_j = \alpha v_j^t + \sum_{i\neq j}W^\text{rec}_{ji}z_i^t + \sum_i W^\text{in}_{ji}x_i^{t+1} - z_j^tv_
            \text{th} \tag{\ref{eq:alifV} revisited}
            \end{equation*}
            and
            \begin{equation*}
            a^{t+1}_j = \rho a^t_j + z^t_j \tag{\ref{eq:alifA} revisited}
            \end{equation*}
            and the update of the observable state

            \begin{equation*}
            z^t_j = H\left(v_j^t - v_\text{th} - \beta a^t_j\right). \tag{\ref{eq:alifZ} revisited}
            \end{equation*}

            The hidden state $\mathbf{h}^t_j$ of an ALIF neuron $j$ is therefore a vector containing its activation and threshold adaptation:
            \begin{equation}
            \mathbf{h}^t_j = \begin{pmatrix}
            v^t_j\\
            a^t_j
            \end{pmatrix}.
            \end{equation}
            This hidden state is associated with a two-dimensional eligibility vector
            \begin{equation}
            \bm{\epsilon}^t_{ji} \eqdef \begin{pmatrix}
            \epsilon_{ji, v}^t\\
            \epsilon_{ji, a}^t
            \end{pmatrix}
            \end{equation}
            that relates to a synapse from any afferent neuron $i$ to neuron $j$.

            Recall from Chapter \ref{ch:introduction} that the eligibility trace slowly fades after a spike has occurred on a synapse, such that a delayed learning signal can still modify the synaptic strength accordingly, solving the credit assignment problem.
            Intuitively, the eligibility vector computes the correct contribution of each of the components of the hidden state.
            For a LIF neuron, the only component is the activation value, and so it is simply a low-pass filter of the spikes of the afferent neuron.

            For the default ALIF neuron, however, the hidden state derivative $\frac{\partial\mathbf{h}^{t+1}_j}{\partial\mathbf{h}^t_j}$ must be computed to derive the eligibility vector.
            This hidden state derivative is expressed by a $2\times2$ matrix of partial hidden state derivatives:
            \begin{equation}
            \frac{\partial\mathbf{h}^{t+1}_j}{\partial\mathbf{h}^t_j} = \begin{pmatrix}
            \frac{\partial v^{t+1}_j}{\partial v^t_j} & \frac{\partial v^{t+1}_j}{\partial a^t_j}\\
            \frac{\partial a^{t+1}_j}{\partial v^t_j} & \frac{\partial a^{t+1}_j}{\partial a^t_j}
            \end{pmatrix}.
            \end{equation}
            The presence of $z^t_j$, and its relation to the Heaviside step function $H(\cdot)$ in the hidden state updates in Equation \ref{eq:alifV} and Equation \ref{eq:alifA} seems problematic for computing these partial derivatives, because the derivative $\frac{\partial z^t_j}{\partial v^t_j}$ is nonexistent.
            This is overcome by replacing it with a simple nonlinear function called a pseudo-derivative.
            Outside of the refractory period of a neuron $j$, this pseudo-derivative has the form
            \begin{equation}
            \psi_j^t = \gamma \max\left(0, 1 - \left|\frac{v_j^t - v_\text{th} - \beta a^t_j}{v_\text{th}}\right|\right),
            \end{equation}
            where $\gamma$ is a dampening constant, which is set to 0 during the neuron's refractory period.
            Like in \citet{esser2016convolutional}, this pseudo-derivative is 1 at time steps where the neuron spikes, and linearly decays to zero in the positive and negative direction.
            The synaptic weight can only change when the pseudo-derivative is nonzero.

            Now, the partial derivatives in the hidden state derivative can be computed by replacing the Heaviside function Equation (in \ref{eq:alifZ}) by the pseudo-derivative $\psi^t_j$:
            \begin{align}
            \frac{\partial v_j^{t+1}}{\partial v_j^t} &= \alpha\\
            \frac{\partial v_j^{t+1}}{\partial a_j^t} &= 0\\
            \frac{\partial a_j^{t+1}}{\partial v_j^t} &= \psi^t_j\\
            \frac{\partial a_j^{t+1}}{\partial a_j^t} &= \rho - \psi^t_j\beta.
            \end{align}
            These partial derivatives can be used to compute the eligibility vector:
            \begin{align}
            \begin{pmatrix}
            \epsilon_{ji, v}^{t+1}\\
            \epsilon_{ji, a}^{t+1}
            \end{pmatrix}
            &=
            \begin{pmatrix}
            \frac{\partial v^{t+1}_j}{\partial v^t_j} & \frac{\partial v^{t+1}_j}{\partial a^t_j}\\
            \frac{\partial a^{t+1}_j}{\partial v^t_j} & \frac{\partial a^{t+1}_j}{\partial a^t_j}
            \end{pmatrix}
            \cdot
            \begin{pmatrix}
            \epsilon_{ji, v}^t\\
            \epsilon_{ji, a}^t
            \end{pmatrix}
            +
            \begin{pmatrix}
            \frac{\partial v^{t+1}_j}{\partial W_{ji}}\\
            \frac{\partial a^{t+1}_j}{\partial W_{ji}}
            \end{pmatrix}\\
            &=
            \begin{pmatrix}
            \alpha & 0\\
            \psi^t_j & \rho-\psi^t_j\beta
            \end{pmatrix}
            \cdot
            \begin{pmatrix}
            \epsilon_{ji, v}^t\\
            \epsilon_{ji, a}^t
            \end{pmatrix}
            +
            \begin{pmatrix}
            z_i^{t-1}\\
            0
            \end{pmatrix}\label{eq:evector_b}\\
            &=
            \begin{pmatrix}
            \alpha \cdot\epsilon_{ji, v}^t + z_i^{t-1}\\
            \psi^t_j\epsilon^t_{ji, v} + \left(\rho-\psi^t_j\beta\right)\epsilon^t_{ji, a}
            \end{pmatrix}.
            \end{align}

            Intuitively, these eligibility vector components can be seen as the contribution of the hidden state component to the increase of the eligibility trace.
            For instance, the activation eligibility component $\epsilon^t_{ji,v}$ of a synapse $i\rightarrow j$ at time step $t$ is, like in the LIF neuron, a low-pass filter of the afferent spikes $z_i$.

            The threshold adaptation eligibility component $\epsilon^t_{ji,a}$ is less intuitive, but acts as a correction factor for the more slowly decaying threshold adaptation.
            Its first term $\psi^t_j\epsilon^t_{ji,v}$ causes it to increase when a neuron has recently spiked and when the activation is already increasing again.
            Therefore, it is higher for synapses that have a higher spike frequency.
            The second term threshold adaptation eligibility component is a decay corrected for the adaptation strength $\beta$.

            This eligibility vector update can be recursively applied.
            For eligibility vectors of synapses that are efferent to input neurons, the input value $x^t_i$ is used in place of $z_i^{t-1}$ in Equation \ref{eq:evector_b}.
            Note that the current time index $t$ is used for input neurons to satisfy the online learning principle defined in the model definition in Equation \ref{eq:model}; neurons receive input from the input at time $t$, and from the spikes of other neurons emitted at time $t-1$.
            Furthermore, the absence of $\epsilon_{ji, a}^t$ in the computation of $\epsilon_{ji, v}^{t+1}$ facilitates online training in emulations in non--von Neumann machines, because $\epsilon_{ji, a}^{t+1}$ can be computed before $\epsilon_{ji, v}^{t+1}$, relieving the need to store a temporary copy of its value.
            In later sections, it is demonstrated that this does not necessarily hold for other neuron models, such as the Izhikevich neuron.

            The eligibility vector needs to be multiplied with the partial derivative of the observable state with respect to the hidden state to obtain the eligibility trace:
            \begin{equation}
            e^t_{ji} = \bm{\epsilon}_{ji}^t \cdot \frac{\partial z^t_j}{\partial\mathbf{h}^t_j}.
            \end{equation}

            Again, the Heaviside function in Equation \ref{eq:alifZ} is replaced by $\psi^t_j$:
            \begin{align}
                \frac{\partial z^t_j}{\partial\mathbf{h}^t_j} &= \begin{pmatrix}
                    \frac{\partial z^t_j}{\partial v^t_j}\\
                    \frac{\partial z^t_j}{\partial a^t_j}
                    \end{pmatrix}\\
                &= \begin{pmatrix}
                    \psi^t_j\\
                    -\beta\psi^t_j
                    \end{pmatrix}.
            \end{align}

            Therefore, the eligibility trace is computed by
            \begin{align}
               e^t_{ji} &= \begin{pmatrix}
            \epsilon_{ji, v}^t\\
            \epsilon_{ji, a}^t
            \end{pmatrix}\cdot \begin{pmatrix}
                    \frac{\partial z^t_j}{\partial v^t_j}\\
                    \frac{\partial z^t_j}{\partial a^t_j}
                    \end{pmatrix}\\
                &= \begin{pmatrix}
            \epsilon_{ji, v}^t\\
            \epsilon_{ji, a}^t
            \end{pmatrix}\cdot \begin{pmatrix}
                    \psi^t_j\\
                    -\beta\psi^t_j
                    \end{pmatrix}\\
                    &= \psi^t_j\left(\epsilon_{ji, v}^t - \beta\epsilon_{ji, a}^t\right).
            \end{align}

\begin{figure}[ht]
                \centering
                \includegraphics[width=\linewidth]{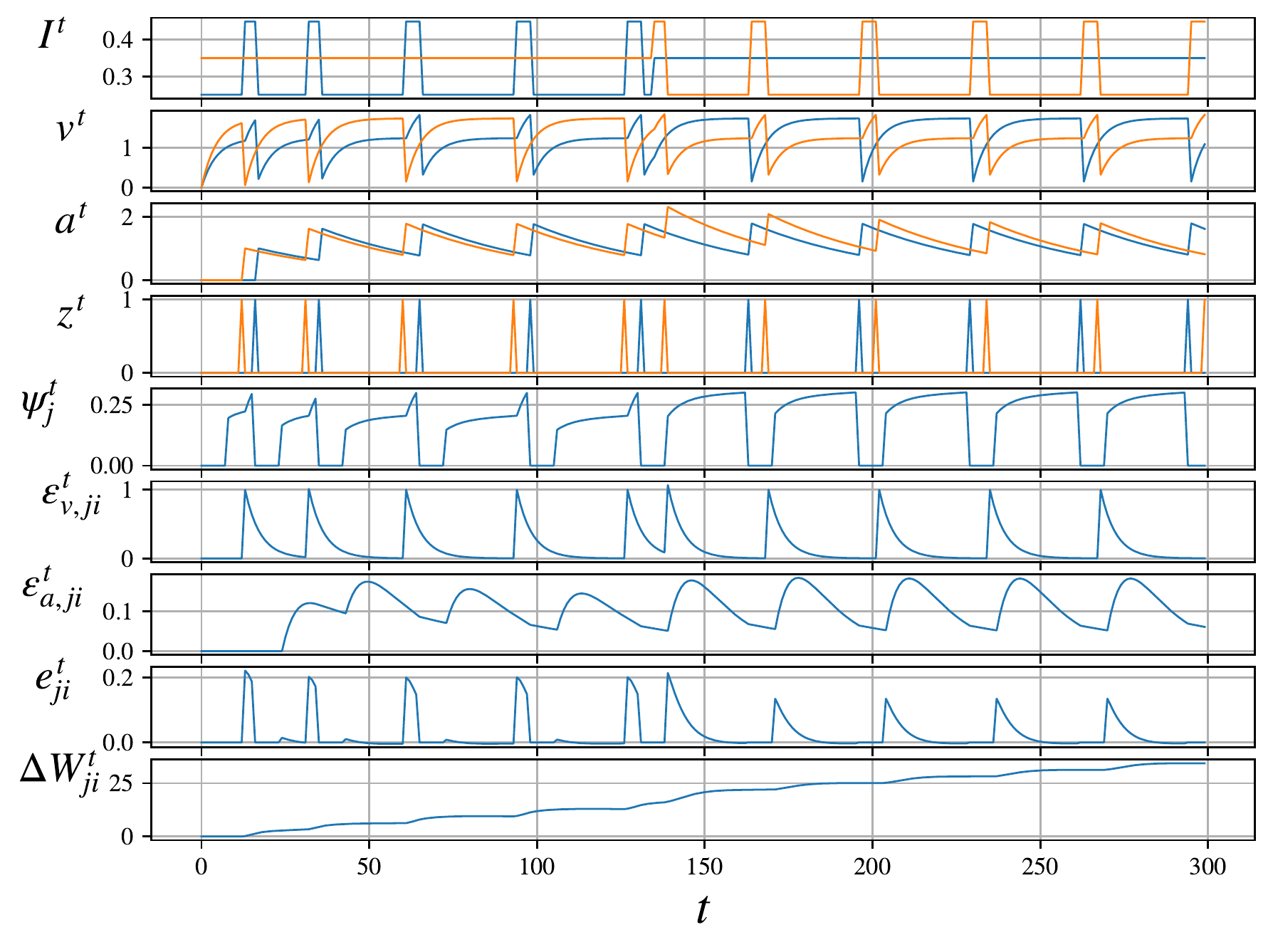}
                \caption[Single-synapse ALIF simulation]{A single-synapse simulation of the evolution of the full hidden state of the ALIF neuron. The blue lines indicate the postsynaptic neuron $j$, and the orange lines indicate the presynaptic neuron $i$. The injected current $I^t$ increases the voltage $v^t_j$ and is deliberately controlled to produce the spike pattern $z^t_j$ where the postsynaptic neuron spikes after the presynaptic neuron during the first half, and vice versa during the second half of the plot. The learning signal $L^t_j$ is kept at a constant value and is omitted for clarity, such that the relation between the eligibility trace $e^t_{ji}$ and the accumulated weight change $\Delta W^t_{ji}$ can be clearly observed. Note that the synapse weight increases regardless of the order of spikes, indicating an absence of STDP in the standard e-prop ALIF neuron.}
                \label{fig:alif}
            \end{figure}

            This means that the eligibility trace can be understood as a low-pass filter of the afferent spikes, with a correction for the efferent neuron's threshold adaptation: a neuron with a higher threshold builds up an eligibility trace more slowly than its more sensitive counterparts.
            Figure \ref{fig:alif} illustrates the behavior of the synaptic variables in an ALIF neuron described above.

        \subsection{Gradients}\label{sec:eprop_grd}
            Gradient descent is used to apply the weight updates, such that weights are updated by a small fraction $\eta$ in the negative direction of the estimated gradient of the loss function with respect to the model weights:
            \begin{equation}\label{eq:eprop_grd}
                \Delta W = -\eta\widehat{\frac{dE}{dW_{ji}}} \eqdef -\eta\sum_t\frac{\partial E}{\partial z^t_j}e^t_{ji}.
            \end{equation}
            Note that for clarity, this section describes e-prop using stochastic gradient descent.
            In the actual implementations in \citet{bellec2020solution} and this research, the Adam optimization algorithm \citep{kingma2014adam} is used (see Section \ref{sec:adam}).

            \paragraph{Error metric}
            In the TIMIT frame-wise phone classification task, there are $K=61$ output neurons $y^t_k$ where $k \in [1\mathrel{{.}\,{.}}\nobreak K]$.
            These are computed according to
            \begin{equation}\label{eq:bellec_y}
            y^t_k = \kappa y^{t-1}_k + \sum_jW^\text{out}_{kj}z^t_j + b_k,
            \end{equation}
            where $\kappa \in [0, 1]$ is the decay factor for the output neurons, $W^\text{out}_{kj}$ is the weight between neuron $j$ and output neuron $k$, and $b_k$ is the bias value.
            The decay factor $\kappa$ acts as a low-pass filter, smoothing the output values over time and implemented based on the observation that output frame classes typically persist for multiple time steps.

            The softmax function $\sigma(\cdot)$ computes the predicted probability $\pi^t_k$ for class $k$ at time $t$:
            \begin{equation}
            \pi^t_k = \sigma_k\left(y^t_1,\ldots,y^t_K\right) = \frac{\exp\left(y^t_k\right)}{\sum_{k'}\exp\left(y^t_{k'}\right)}.
            \end{equation}
            This predicted probability is compared with the one-hot vector corresponding to the target class label $\pi^{*,t}_k$ at time step $t$ using the cross entropy loss function
            \begin{equation}
            E = -\sum_{t,k}\pi^{*,t}_k\log\pi^t_k,
            \end{equation}
            thereby obtaining the accumulated loss $E$ at time step $t$.

            Since the learning signal $L^t_j$ is defined as the partial derivative of the error $E$ with respect to the observable state $z_j^t$ of a neuron $j$ afferent to an output neuron $k$, we can use
            \begin{equation}\label{eq:learningsignal_after_output}
            L^t_j = \frac{\partial E}{\partial z^t_j} = \sum_kB_{jk}\sum_{t'\geq t}\left(\pi^{t'}_k - \pi^{*,t'}_k\right)\kappa^{t'-t},
            \end{equation}
            where $B_{jk}$ is a feedback weight from neuron $k$ back to output neuron $j$.
            There are multiple strategies for choosing feedback weights.
            \citet{bellec2020solution} noted that a constantly uniform weight matrix yields poor performance, which has been empirically verified in my project.
            However, when the feedback weight matrix is initialized from a zero-centered normal distribution, it can remain either constant, mirror $(W^\text{out})^\top$, or update according to $(\Delta W^\text{out})^\top$.
            These three variants are referred to in \citet{bellec2020solution} as \emph{random}, \emph{symmetric}, and \emph{adaptive} e-prop, respectively.
            In this paper, symmetric e-prop is used (i.e., $B_{jk} \eqdef W^\text{out}_{kj}$) unless explicitly stated otherwise.

            Note that the term $\kappa^{t'-t}$ in Equation \ref{eq:learningsignal_after_output} is a filter that compensates for the decay factor of output neurons.
            Note also that this equation does not allow online learning, because future time steps $t'$ are accessed.
            However, if a low-pass filter with factor $\kappa$ is applied on the eligibility trace, it will cancel out the effects of the future time steps on the learning signal, and the estimated loss gradient can be approximated.
            This low-pass filter of the eligibility trace can be implemented in an online fashion by including it as a hidden synaptic variable $\bar{e}^t_{ji}$.
            Recall that the estimated loss gradient $\widehat{\frac{dE}{dW_{ji}}}$ is approximated by $\sum_t \frac{\partial E}{\partial z^t_j}e^t_{ji}$.
            Therefore, after inserting Equation \ref{eq:learningsignal_after_output} in Equation \ref{eq:eprop_grd}, the weight update is computed by
            \begin{align}
            \Delta W_{ji} &= -\eta\sum_{t'}\frac{\partial E}{\partial z^{t'}_j}e^{t'}_{ji}\\
            &= -\eta\sum_{t'}\sum_kB_{jk}\sum_{t'\geq t}\left(\pi^{t'}_k - \pi^{*,t}_k\right)\kappa^{t'-t}e^{t'}_{ji}\\
            &= -\eta\sum_{k, t'}B_{jk}\sum_{t'\geq t}\left(\pi^{t'}_k - \pi^{*,t}_k\right)\kappa^{t'-t}e^{t'}_{ji}\\
            &= -\eta\sum_t\underbrace{\sum_kB_{jk}\left(\pi^{t}_k - \pi^{*,t}_k\right)}_{=L^t_j}\underbrace{\sum_{t'\leq t}\kappa^{t'-t}e^{t'}_{ji}}_{\eqdef \bar{e}^t_{ji}}\label{dwlast},
            \end{align}
            where $W_{ji}$ is an input or recurrent weight.
            By implementing $\bar{e}_{ji}$ as a low-pass filter (with factor $\kappa$) of the eligibility trace, the weight update in Equation \ref{dwlast} is implemented as a local and online learning algorithm.

            The training algorithm for the output weights $W^\text{out}$ and bias $b$ can be directly derived from gradient descent:
            \begin{equation}
            \Delta W^\text{out}_{kj} = -\eta \sum_t\left(\pi^t_k - \pi^{*,t}_k\right)\sum_{t'\leq t}\kappa^{t'-t}z^t_j
            \end{equation}
            and
            \begin{equation}
            \Delta b_k = -\eta \sum_t\left(\pi^t_k - \pi^{*,t}_k\right).
            \end{equation}

%% file: Chapters/Chapter03.tex
\chapter{Method}\label{ch:method}
\section{Data Preprocessing}

	In this section, the content of the TIMIT speech corpus is described, as well as the preprocessing method that transforms the speech signals into usable features in the e-prop framework.

	\subsection{The TIMIT speech corpus}
		TIMIT is a speech corpus that contains phonemically transcribed speech~\citep{garofolo1993darpa}, comprising 6300 sentences, 10 spoken by each of the 630 speakers.
		To include a broad range of dialects these speakers are sampled from 8 different geographical regions in the United States (as categorized in \citet{labov2008atlas}) in which they lived during their childhood years.
		Table~\ref{tab:dialects} breaks down the precise composition of the dialect distribution.

		\begin{table}[ht]
		    \myfloatalign
		    \begin{tabularx}{\textwidth}{lrrr} \toprule
		        \tableheadline{Dialect region} & \tableheadline{\#Male}
		        & \tableheadline{\#Female} & \tableheadline{Total} \\ \midrule
		        1 (New England)   & 31 (63\%) & 18 (27\%) &  49  \phantom{0}(8\%)  \\
		        2 (Northern)      & 71 (70\%) & 31 (30\%) & 102 (16\%) \\
		        3 (North Midland) & 79 (67\%) & 23 (23\%) & 102 (16\%) \\
		        4 (South Midland) & 69 (69\%) & 31 (31\%) & 100 (16\%) \\
		        5 (Southern)      & 62 (63\%) & 36 (37\%) &  98 (16\%) \\
		        6 (New York City) & 30 (65\%) & 16 (35\%) &  46  \phantom{0}(7\%)  \\
		        7 (Western)       & 74 (74\%) & 26 (26\%) & 100 (16\%) \\
		        8                 & 22 (67\%) & 11 (33\%) &  33  \phantom{0}(5\%)  \\
		        \midrule
		        All  & 438 (70\%) & 192 (30\%) & 630 (100\%) \\
		        \bottomrule
		    \end{tabularx}
		    \caption[TIMIT dialect regions]{Distribution of speakers' dialect regions and sexes. Speakers of the innominate dialect region 8 relocated often during their childhood.}  \label{tab:dialects}
		\end{table}

		The sentence text can be categorized into 2 \emph{dialect} sentences, 450 \emph{phonetically compact} sentences, and 1890 \emph{phonetically diverse} sentences.

		The dialect sentences, which are spoken by all speakers, are designed to expose the dialectical variants of the speakers.
		The phonetically compact sentences are designed to include many pairs of phones.
		The phonetically diverse sentences are taken from the Brown Corpus~\citep{kucera1967computational} and the Playwrights Dialog~\citep{hultzsch1964tables} in order to maximize the number of allophones (\ie, different phones used to pronounce the same phoneme).
		Table~\ref{tab:types} lists an overview of the distribution of the number of speakers per sentence type.

		\begin{table}[ht]
		    \myfloatalign
		    \begin{tabularx}{\textwidth}{lrrrr} \toprule
		        \tableheadline{Sentence type} & \tableheadline{\#Sentences}
		        & \tableheadline{\#Speakers} & \tableheadline{Total} \\ \midrule
		        Dialect & 2    & 630 & 1260\\
		        Compact & 450  & 7   & 3150 \\
		        Diverse & 1890 & 1   & 1890 \\
		        \midrule
		        Total   & 2342 &     & 6300 \\
		        \bottomrule
		    \end{tabularx}
		    \caption[TIMIT sentence types]{Distribution of sentence types.}
		    \label{tab:types}
		\end{table}

		Each of the sentences is encoded in as a waveform signal in \texttt{.wav} format, and is accompanied by a corresponding text file indicating which phones are pronounced in the waveform, and between which pairs of sample points.

	\subsection{Data splitting}
		The TIMIT dataset is split into a training, validation and testing set as in \citet{graves2005framewise} and \citet{bellec2020solution}.
		The training set is used to train the network synaptic weights according to the e-prop algorithm.
		The validation set is used to obtain a well-performing set of hyperparameters.
		The testing set is used to evaluate the performance of the network after the hyperparameters are obtained.

		The TIMIT corpus documentation offers a suggested partitioning of the training and testing data, which is based on the following criteria:
		\begin{enumerate}
			\item 70\%--80\% of the data is used for training, and the remaining 20\%--30\% for testing.
			\item No speaker appears in both the training and testing partitions.
			\item Both subsets include at least 1 male and 1 female speaker from every dialect region.
			\item There is a minimal overlap of text material in the two subsets.
			\item The test set should contain all phones in as many allophonic contexts as possible.
		\end{enumerate}
		In accordance with these criteria, the TIMIT corpus includes a ``core'' test set that contains 2 male speakers and 1 female speaker from each dialect, summing up to 24 speakers.
		Each of these speakers read a different set of 5 phonetically compact sentences, and 3 phonetically diverse sentences that were unique for each speaker.
		Consequently, the test set comprises 192 sentences ($24\times(5+3)$) and was selected such that it contains at least one occurrence of each phone.
		In this report, the TIMIT core test set is used, thereby meeting the criteria listed above.

		The remaining 4096 sentences are randomly partitioned into 3696 training sentences and 400 validation sentences in this research (the TIMIT corpus contains no fixed training/validation set partition).

	\subsection{Engineering features}

\begin{figure}[ht]
				\centering
			    \includegraphics[width=\linewidth]{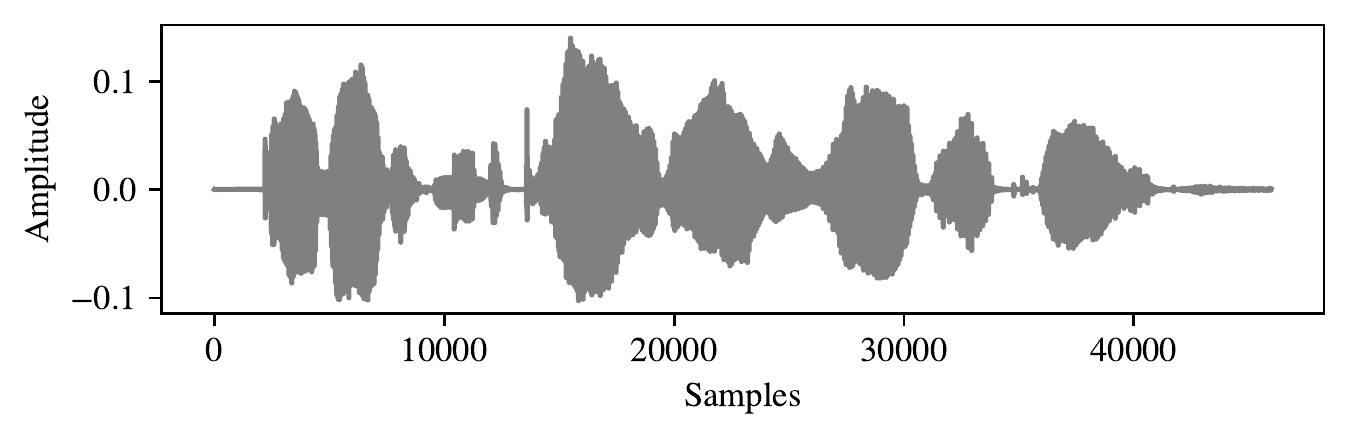}
			    \caption{Raw TIMIT waveform signal}
			    \label{fig:signal}
			\end{figure}

		The data preprocessing pipeline is similar to that used in \citet{fayek2016}, which can be summarized by applying a pre-emphasis filter on the waveforms, then slicing the waveform in short frames, taking their short-term power spectra, computing 26 filterbanks, and finally obtain 12 Mel-Frequency Cepstrum Coefficients (MFCCs).
		Then, these MFCCs are aligned with the phones found in the TIMIT dataset.
		An example of a waveform signal is given in Figure \ref{fig:signal}.

		\paragraph{Pre-emphasis}

\begin{figure}[ht]
				\centering
			    \includegraphics[width=\linewidth]{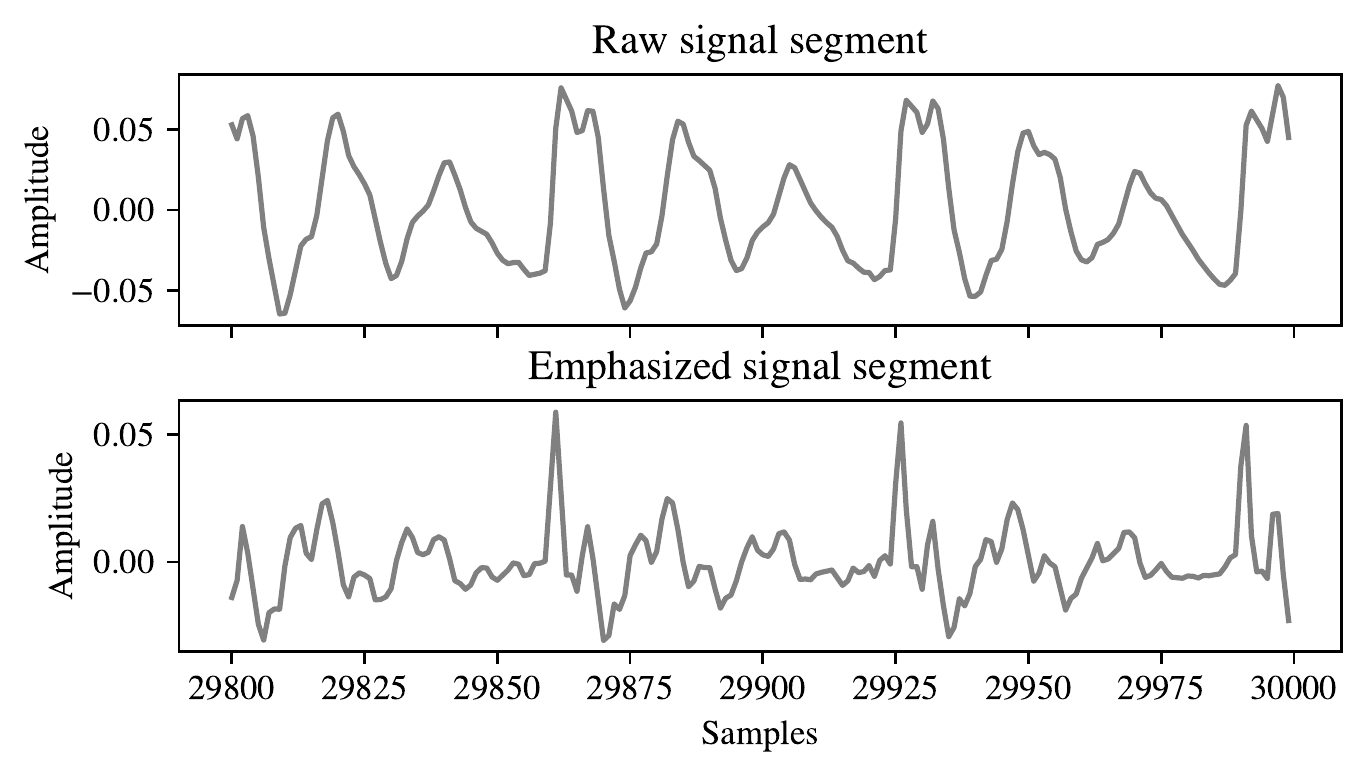}
			    \caption[Pre-emphasis filtered signal segment]{A segment of a signal after the pre-emphasis filter of Equation \ref{eq:pre_emphasis} was applied to it. The upper panel contains samples 28000--30000 of the signal in Figure \ref{fig:signal}, while the lower panel contains its pre-emphasis filtered counterpart. Note that the filtered signal is less symmetric around the horizontal axis, because the filtered signal is similar to a first derivative---the original signal has sharper increases than decreases, so the filtered signal has stronger extrema towards the positive direction.}
			    \label{fig:signalemph}
			\end{figure}

			In speech signals, high frequencies generally have smaller magnitudes than lower frequencies.
			To balance the magnitudes over the range of frequencies in the signal, a pre-emphasis filter $y(t)$ is applied on the waveform signal $x(t)$:
			\begin{equation}\label{eq:pre_emphasis}
				y(t) = x(t) - 0.97x(t-1).
			\end{equation}
			This procedure yields the additional benefit of improving the signal-to-noise ratio.
			An example of a pre-emphasized signal is given in Figure \ref{fig:signalemph}.

		\paragraph{Framing}
			The waveforms, which are sampled at a rate $f_s$ of \SI{16}{\kHz}, cannot be directly used as input to the model, because they are too long---a typical sentence waveform contains in the order of tens of thousands of data points.
			Furthermore, the individual data points are not very informative, because they reflect the sound wave of the uttered sound, not the characteristics of the source of this sound.
			These sounds are filtered by the shape of the vocal tract, which manifests itself in the envelope of the short time power spectrum of the sound.
			This power spectrum representation describes the power of the frequency components of the signal over a brief interval.
			The frequency components are assumed to be stationary over short intervals, in contrast to the full sentence, which carries its meaning because it is non-stationary.
			Therefore, the waveform signals are transformed into series of frequency coefficients of short-term power spectra.
			To obtain these multiple short-term power spectra over the duration of the waveform, it is sliced into brief overlapping frames.

			Every 160 samples (equivalent to \SI{10}{ms}) of a pre-emphasized signal, an interval frame of 400 samples (equivalent to \SI{25}{ms}) is extracted.
			This means that the frames overlap by \SI{25}{ms}.
			The waveform is zero-padded such that the last frame also has 400 samples.
			By this process, signal frames $x_i(n)$ are obtained, where $n$ ranges over 1--400, and $i$ ranges over the number of frames in the waveform.

\begin{figure}[ht]
				\centering
			    \includegraphics[width=0.45\linewidth]{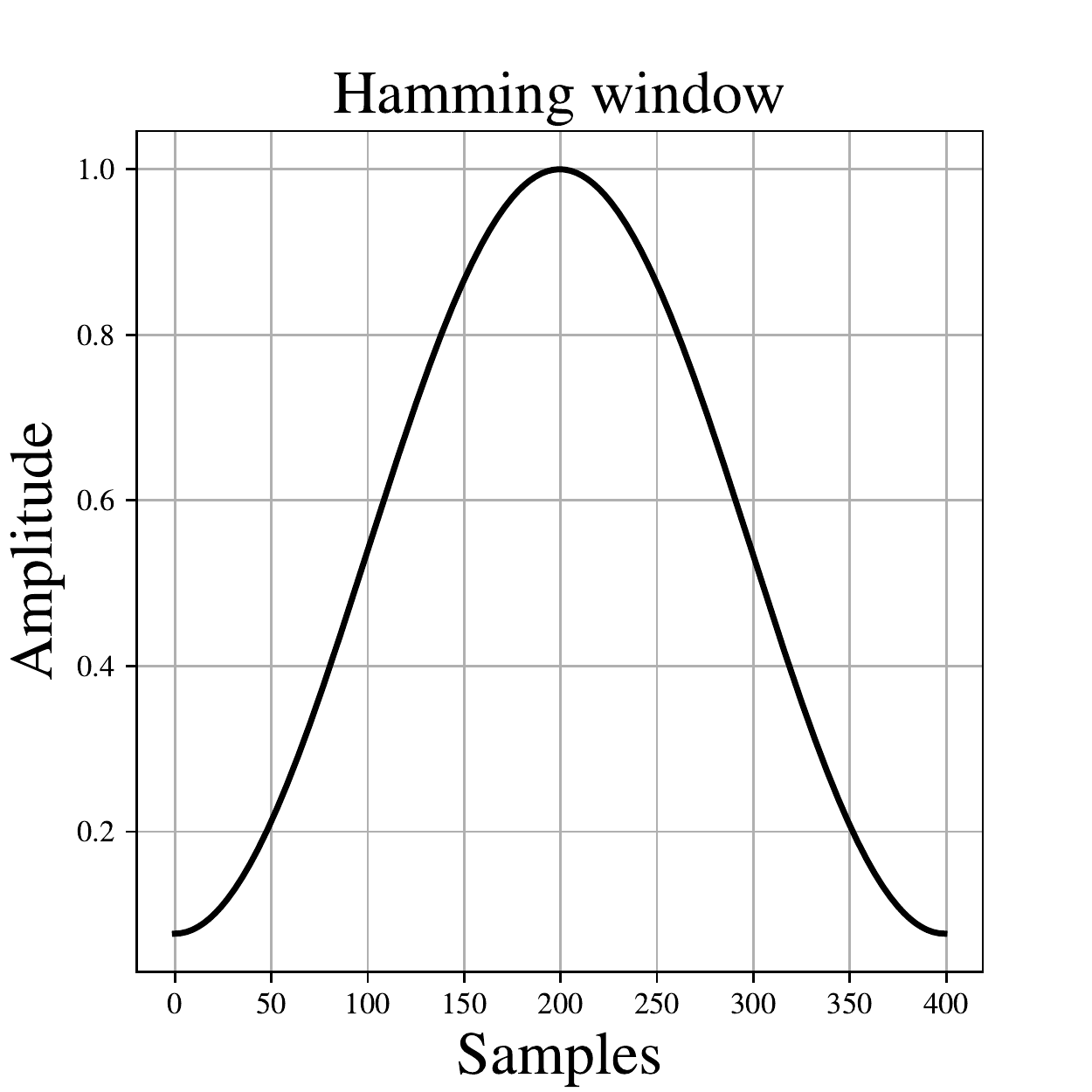}
			    \caption[Hamming window]{The form of the Hamming window applied on signal frames to reduce spectral leakage.}
			    \label{fig:hamming}
			\end{figure}

			Then, a Hamming window with the form
			\begin{equation}\label{eq:hamming}
				w\left[n\right] = a_0 - a_1\cos\left(\frac{2\pi n}{N-1}\right),
			\end{equation}
			is applied where $N$ is the window length of 400 samples, $0 \leq n < N$, $a_0 = 0.53836$, and $a_1 = 0.46164$.
			A plot of this window is given in Figure \ref{fig:hamming}.
			This window is applied to reduce the spectral leakage, which manifests itself though sidelobes in the power spectra.
			Applying the Hamming window reduces the sidelobes to near-equiripple conditions, minimizing the leakage \citep{SASPWEB2011}.

		\paragraph{Short-term power spectra}

			The power spectra $P_i$ are obtained for each frame by first taking the absolute $K$-point discrete Fourier transform (DFT) of the frame samples $x_i(n)$:
			\begin{equation}
				X_k = \left|\sum_{n=0}^{N-1}x_i(n)\cdot e^{-\frac{i2\pi}{N}kn}\right|,
			\end{equation}
			where $K=512$.
			This yields the magnitudes of the discrete cosine transform (DCT) of the frames.

\begin{figure}[ht]
				\centering
			    \includegraphics[width=0.45\linewidth]{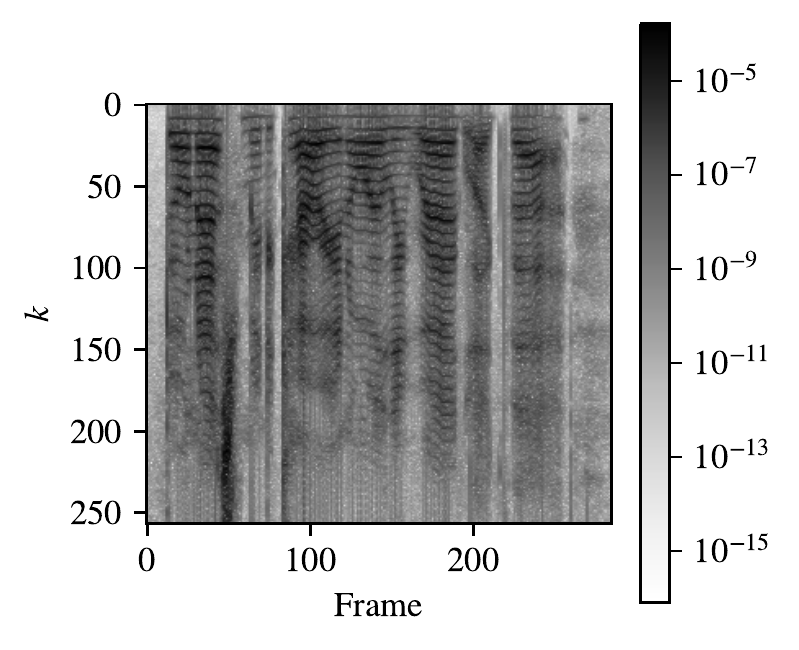}
			    \caption[Power spectra]{The power spectra of a sentence.}
			    \label{fig:powframes}
			\end{figure}

			The power spectra are obtained using the equation
			\begin{equation}\label{eq:powframes}
				P = \frac{{X_k}^2}{K},
			\end{equation}
			an example of which is shown in Figure \ref{fig:powframes}.

		\paragraph{Mel filterbank}

			The short-term power spectra are then transformed to Mel-spaced filterbanks.
			The Mel scale is a scale of pitches that are perceptually equal in distance \citep{stevens1937scale}.
			This is in contrast to the frequency measurement, in which the human cochlea can distinguish lower frequencies more accurately than higher ones.
			The aim of converting to the Mel scale is to make every filterbank coefficient feature equally informative, thereby improving the learning performance of the model.

			The Mel-spaced filterbank is a set of 40 triangular filters that we apply to each frame in $P$.

			To compute the Mel-spaced filterbank, lower and upper band edges of respectively \SI{0}{\Hz} and $f_s/2 = \SI{8}{\kHz}$ are selected, and convert these to Mels using
			\begin{equation}
				m(f) = 2595\log_{10}\left(1 + \frac{f}{700}\right),
			\end{equation}
			where $f$ is the frequency in $\SI{}{\Hz}$.
			This yields a lower band edge of 0 Mels and an upper band edge of approximately 2835 Mels.

			The 40 filterbanks are obtained by first spacing 42 points $\mathbf{m}$ linearly between these bounds (inclusive), thereby obtaining 40 points spaced exclusively between the bounds.

			Then, the vector of Mel frequencies $\mathbf{m}$ is converted back to \SI{}{\Hz} using
			\begin{equation}
				\mathbf{f} = 700\left(10^{\mathbf{m}/2595}-1\right).
			\end{equation}
			The resulting frequencies $\mathbf{f}$ are rounded to their nearest Fourier transform bins $\mathbf{b}$ using
			\begin{equation}
				\mathbf{b} = \lfloor(K+1)\mathbf{f}/fs\rfloor.
			\end{equation}

			The resulting 40 filterbanks with their corresponding Mels and frequencies are listed in Table \ref{tab:mels}.

\begin{figure}[ht]
				\centering
			    \includegraphics[width=\linewidth]{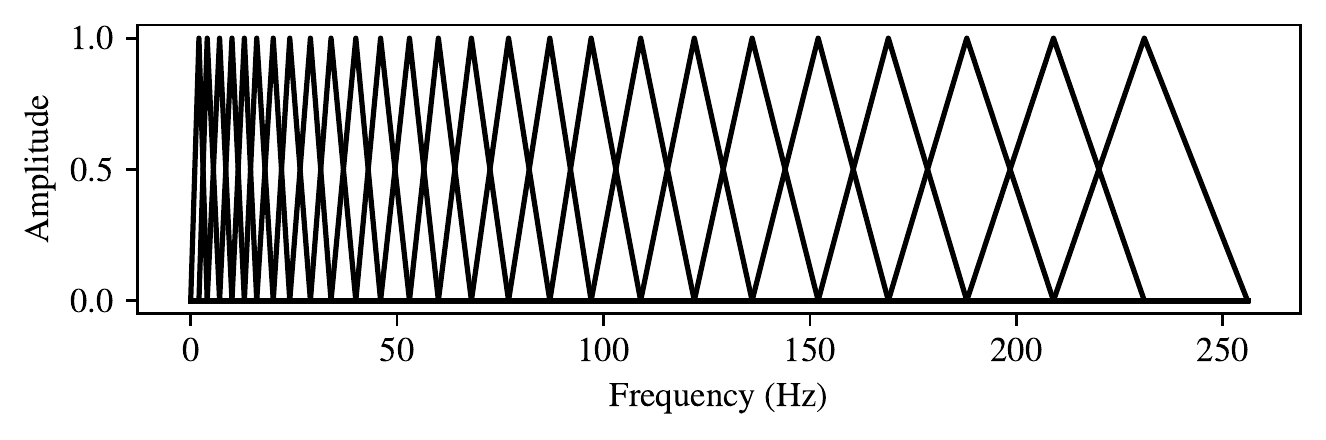}
			    \caption[Mel-spaced filterbanks]{The Mel-spaced filterbanks.}
			    \label{fig:filterbank}
			\end{figure}

			The $i\textsuperscript{th}$ filter in filterbank $H_i$ is a triangular filter that has its lower boundary at $b_{i}$ \SI{}{\Hz}, its peak at $b_{i+1}$ \SI{}{\Hz}, and its upper boundary at $b_{i+2}$ \SI{}{\Hz}.
			For other frequencies, they are 0.
			Therefore, the filterbank can be described by
			\begin{equation}
				H_i(k) = \begin{cases}
					0 & \mbox{if } k<b_i\\
					\frac{k-b_i}{b_{i+1}-b_i} & \mbox{if } b_i\leq k < b_{i+1} \\
					1 & \mbox{if } k = b_{i+1} \\
					\frac{b_{i+2} - k}{b_{i+2}-b_{i+1}} & \mbox{if } b_{i+1} < k \leq b_{i+2}\\
					0 & \mbox{if } b_{i+2} < k
				\end{cases},
			\end{equation}
			where $0 \leq k \leq \frac{K}{2}$.
			These Mel-spaced filters are shown in Figure \ref{fig:filterbank}.

\begin{figure}[ht]
				\centering
			    \includegraphics[width=\linewidth]{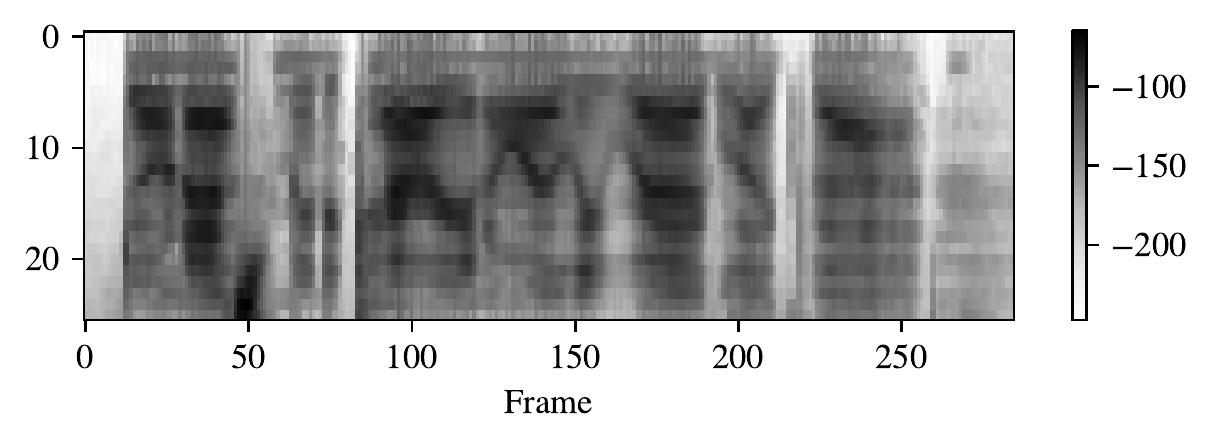}
			    \caption[Spectrogram]{An example of the spectrogram of a sentence.}
			    \label{fig:spectrogram}
			\end{figure}

			After applying the filterbank to the short-term power spectrum, a spectrogram $S$ of the frame sequence (see \eg~Figure \ref{fig:spectrogram}) is obtained.

		\paragraph{Mel-frequency cepstral coefficients}

			Coefficients in the spectrograms are strongly correlated, which would negatively impact the learning performance of the model.
			Therefore, the DCT is applied again to decorrelate the coefficients and obtain the power cepstrum $C$ of the speech frame:

			\begin{equation}
				C_k = 2c\sum^{N-1}_{n=0}S\left(n\right)\cos\left(\frac{\pi k\left(2n+1\right)}{2N}\right),
			\end{equation}
			where $c$ is a scaling factor that makes the matrix of coefficients orthonormal:
			\begin{equation}
			c = \begin{cases}
				\sqrt{\frac{1}{4N}} & \mbox{if } k = 0,\\
				\sqrt{\frac{1}{2N}} & \mbox{otherwise.}
			\end{cases}
			\end{equation}

			The first coefficient in $C$ is discarded, because it indicates the average power of the input signal and therefore carries little meaning.
			Coefficients higher than 13 are also discarded, because they represent only fast changes in the spectrogram and increase the complexity of the input signal while adding increasingly less meaning to it.
			Next, the first and second derivatives of the MFCCs over time are concatenated to the 13 MFCC features, obtaining an input vector of size 39.

\begin{figure}[!ht]
			    \myfloatalign
			    \subfloat[MFCCs centered around 0 per channel. Note that these MFCCs are standardized channel-wise over the full training dataset, and that the first and second derivations of the MFCC features are omitted.]
			    {\includegraphics[width=\linewidth]{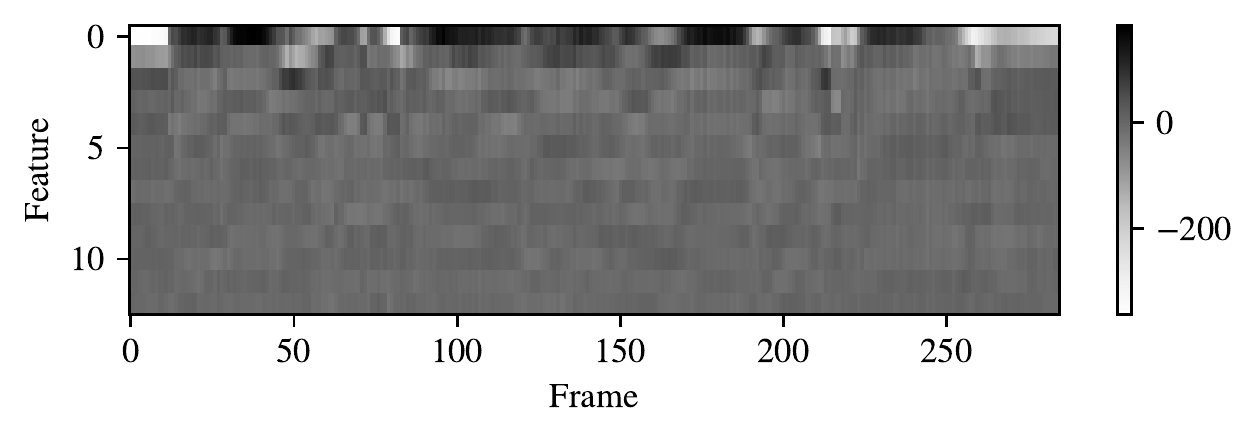}} \\
			    \subfloat[Target signal encoded as a one-hot vector changing over time. The order of phones along the one-hot vector corresponds to the order in which they are encountered in processing the dataset. This particular example shows a pattern because it is the first processed sentence in the training set.]
			    {\includegraphics[width=\linewidth]{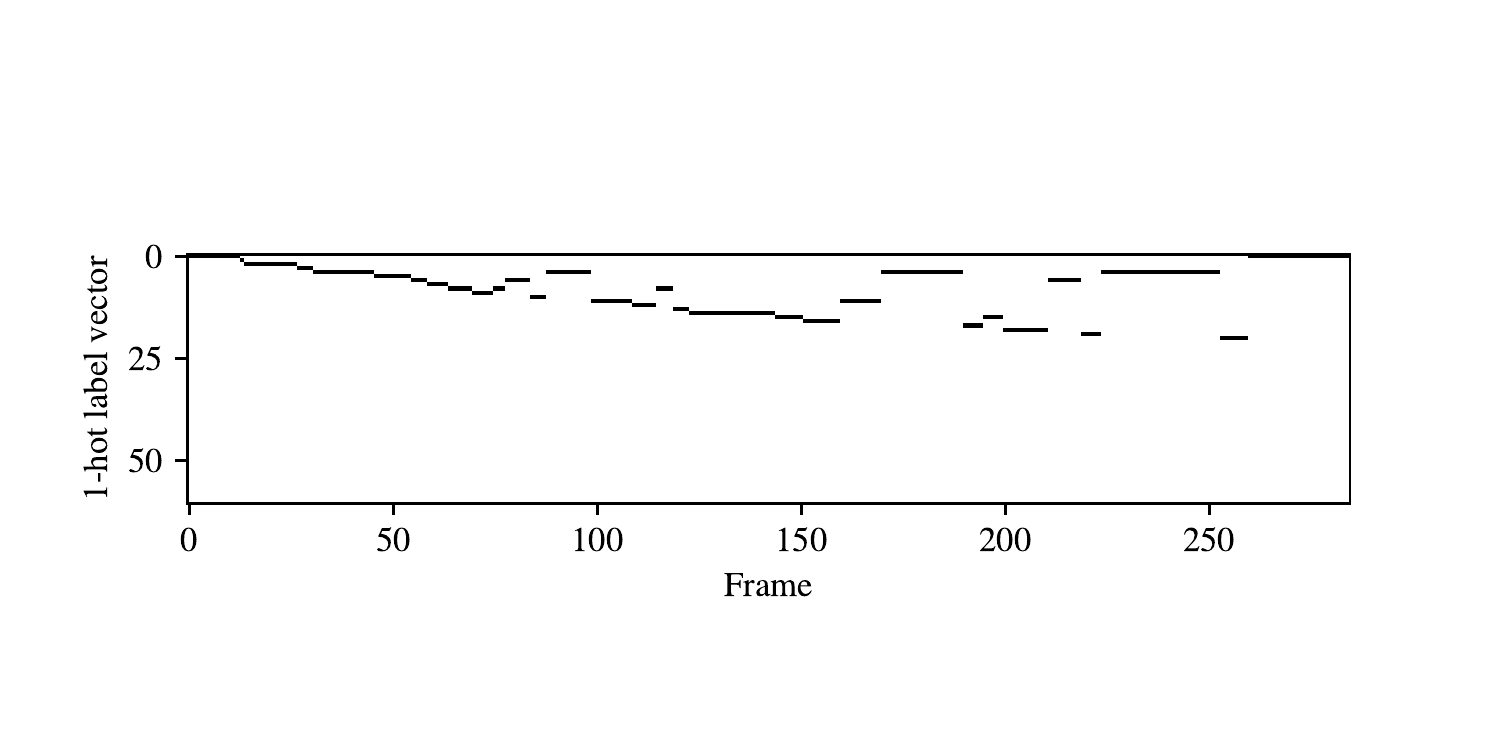}}
			    \caption[Input/features/targets]{MFCCs features and corresponding target phones.}\label{fig:source_mfcc_target}
			\end{figure}

			Then, these input vectors are balanced by centering each frame around the value 0.
			An example of these input vectors is given in Figure \ref{fig:source_mfcc_target}.
			Note that before training a model, all input vectors (including validation and testing data) are standardized channel-wise according to the full training set (see \ref{fig:inoutpair} for an example input as used by the model).

		\paragraph{Target output}

			The target output of the model is a frame-wise representation of the phones that are uttered in a sentence.
			The TIMIT corpus contains text files indicating in what order phones occur in a sentence, and their starting and ending sample points.

			These phones are discretized into frames such that they align correctly with the MFCCs.
			They are represented in one-hot vector encoding.
			Since the dataset contains 61 different phones, this is also the length of these vectors.

			Figure \ref{fig:source_mfcc_target} illustrates the waveform data and its frame-wise aligned MFCCs and target output.
			Note that the full dataset of features is standardized per training data channel; feature channels are first centered around 0, and then divided by their standard deviations.
			To prevent data leakage, validation and testing data are standardized according to the means and standard deviations in the training data.
			An example of a standardized input is shown in Figure \ref{fig:inoutpair}.

\newpage

\section{Enhancing e-prop}
	In my project I examined two types of enhancements to apply the e-prop learning algorithm on the TIMIT dataset.

	The first type is the effect of the neuron model; particularly, the effect of including STDP behavior is analyzed.
	The second type is the effect of a multi-layered architecture.

	The results of these enhancements will answer the research questions posed in Chapter \ref{ch:introduction}, i.e., whether multi-layered architectures or inclusion of STDP in neuron models improves the performance of e-prop.

	\subsection{Multi-layer architecture}\label{sec:ml_arch}

		\begin{figure}[!ht]
		    \myfloatalign
		    \includegraphics[trim=0 25cm 0 0, clip, width=\linewidth]{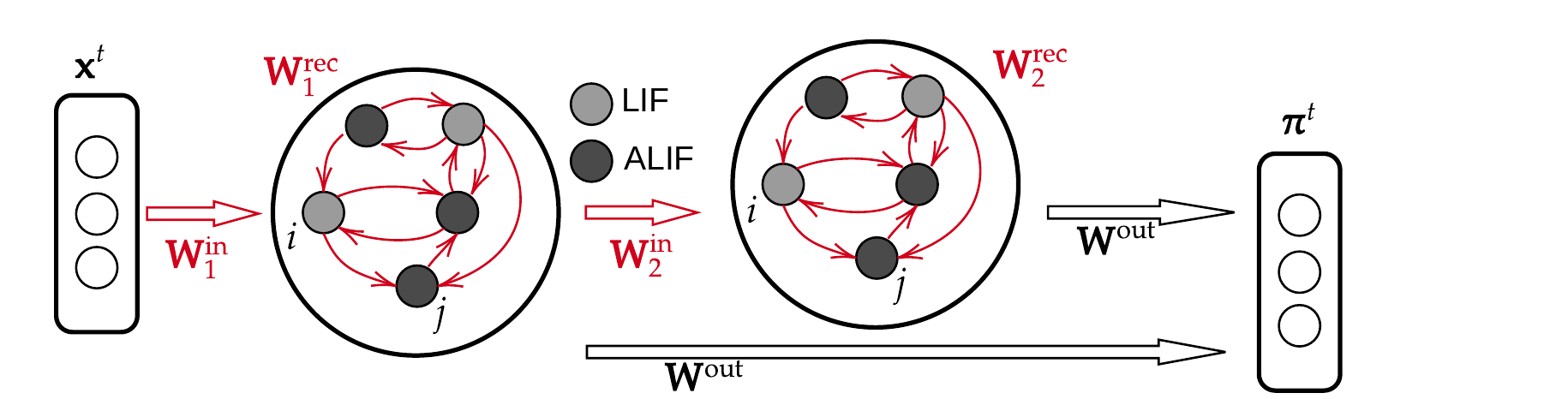}
		    \caption[Multi-layer illustration]{An illustration of a multi-layer network architecture. Some details shown in Figure \ref{fig:topology-sl} are omitted for clarity.}
		    \label{fig:topology-ml}
		  \end{figure}

		The multi-layer e-prop architecture can be described in the same formal model as its single-layer counterpart, in which the hidden state is based on temporally (i.e., online) and spatially locally available information at a neuron $j$:

        \begin{equation}
        \mathbf{h}^t_j = M\left(\mathbf{h}_j^{t-1}, \mathbf{z}^{t-1}, \mathbf{x}^t, \mathbf{W}_j\right).\tag{\ref{eq:model} revisited}
        \end{equation}
        For the multi-layer architecture, however, neurons in deeper layers no longer depend on the input, but on the observable states of the previous layer at the same time step, such that at every time step, a full pass through the network is made.
        We modify the indexing notation accordingly, in order to directly refer to neurons and weights in a particular layer $r \in [1\mathrel{{.}\,{.}}\nobreak R]$:
        \begin{equation}\label{eq:ml_model}
        \mathbf{h}^t_{rj} = \begin{cases}
        M\left(\mathbf{h}_{rj}^{t-1}, \mathbf{z}_r^{t-1}, \mathbf{x}^t, \mathbf{W}_{rj}\right)       & \mbox{if } r = 1\\
        M\left(\mathbf{h}_{rj}^{t-1}, \mathbf{z}_r^{t-1}, \mathbf{z}_{r-1}^t, \mathbf{W}_{rj}\right) & \mbox{otherwise,}
        \end{cases}
        \end{equation}
        where $\mathbf{h}^t_{rj}$ (resp. $z^t_{rj}$) is the hidden state (resp. observable state) of a neuron $j$ in layer $r$, and $\mathbf{W}_{rj} = \mathbf{W}^\text{in}_{rj} \cup \mathbf{W}^\text{rec}_{rj}$ is the set of afferent weights to neuron $j$ in layer $r$.

        Similarly, the observable state can be modeled by
        \begin{equation}\label{eq:ml_model_obs}
        z^t_{rj} = f\left(\mathbf{h}_{rj}^t\right)
        \end{equation}
        and the network output by
        \begin{equation}
        y^t_k = \kappa y^{t-1}_k + \sum_{j,r}W^\text{out}_{rkj}z_{rj}^t + b_k,
        \end{equation}
        where $W^\text{out}_{rkj}$ is a weight between neuron $j$ in layer $r$ and output neuron $k$.
        Note that the summation over $r$ entails that the output layer is connected to all neurons in all layers in the network.
        This allows trainable broadcast weights in earlier layers, such as those found in symmetric and adaptive e-prop.

        \paragraph{Multi-layer ALIF neurons}
        An ALIF neuron in a multi-layer architecture is similar to one in a single-layer architecture (see Section \ref{sec:alif}).
        The only difference, apart from the layer indexing, is its activity update.
        For a multi-layer ALIF neuron, the activity value is given by
        \begin{equation}\label{eq:ml_alifV}
        v^{t+1}_{rj} = \alpha v_{rj}^t + \sum_{i\neq j}W^\text{rec}_{rji}z_i^t + \sum_i W^\text{in}_{rji}I - z_{rj}^tv_
        \text{th},
        \end{equation}
        where
        \begin{equation}\label{eq:inpVml}
        I = \begin{cases}
        	x^{t+1}_i       &\mbox{if } r = 1 \\
            z^{t+1}_{r-1,i} &\mbox{otherwise.}
            \end{cases}
        \end{equation}

	\subsection{Other neuron types}

		In this section, the STDP-ALIF and Izhikevich neuron models are presented.
		The advantage of these models over the standard ALIF model is that they naturally elicit STDP.
		Additionally, the Izhikevich model has an implicit refractory mechanism that is built into its system of equations, making it a more biologically plausible model, as opposed to the (STDP-)ALIF model that requires an explicit timer variable in a neuron's hidden state (but it is still local and online).

		The Izhikevich e-prop neuron model was first presented by \citet{traub2020learning} only in a single-synapse demonstration of its STDP properties.
		In this report, the performance of the e-prop Izhikevich model is experimentally validated in a learning task for the first time.
		\citet{traub2020learning} also described the STDP-LIF, which is a non-adaptive modification of the standard LIF neuron.
		Here, its adaptive counterpart, the STDP-ALIF model, is derived and validated as well.
		This allows for a direct comparison between the ALIF and STDP-ALIF models, such that the effects of STDP can be more precisely analyzed.

		\subsubsection{STDP-ALIF}
			The key change between the ALIF and STDP-ALIF neuron is that the latter is reset to zero at a spike event, and when its refractory period of $\delta t_\text{ref}$ time steps ends.
			Recall that in contrast, the standard ALIF neuron only uses a soft reset by including a term $-z^t_{rj}v_\text{th}$ in its activation update equation (Equation \ref{eq:ml_alifV}).

			The activation update of the STDP-ALIF neuron is therefore:
			\begin{equation}\label{eq:ml_stdpalifV}
	        v^{t+1}_{rj} = \alpha v_{rj}^t + \sum_{i\neq j}W^\text{rec}_{rji}z_i^t + \sum_i W^\text{in}_{rji}I -z^t_{rj}\alpha v^t_{rj} - z_{rj}^{t-\delta t_\text{ref}}\alpha v^t_{rj},
	        \end{equation}
	        where, again, $I$ is the network input if $r=1$, otherwise it is the observable state of neuron $i$ in the preceding layer (see Equation \ref{eq:inpVml}).
	        Recall that a neuron cannot spike for $T^\text{refr}$ time steps after its last spike---therefore, the neuron is still suppressed at time step $t-\delta t_\text{ref}$ and hence the fourth and fifth terms $-z^t_{rj}\alpha v^t_{rj}$ and $- z_{rj}^{t-\delta t_\text{ref}}\alpha v^t_{rj}$ cannot simultaneously be nonzero.
	        Note also that Equation \ref{eq:ml_stdpalifV} will not necessarily set the activation value precisely to 0 at spike time and after the refractory time, as only the activation that caused the neuron to spike will be subtracted, and the new input values described in its second and third term will be added to the new activation value.

	        The hidden state derivative changes accordingly:
	        \begin{align}
	        \frac{\partial v_{rj}^{t+1}}{\partial v^t_{rj}} &= \alpha - z^t_{rj}\alpha - \alpha z_{rj}^{t-\delta t_\text{ref}}\\
	        &= \alpha\left(1 - z^t_{rj} - z_{rj}^{t-\delta t_\text{ref}}\right).
	        \end{align}
	        Note that the absence of $a^t_{rj}$ in the new activation update entails that the other entries of the hidden state Jacobian are equal to those of the ALIF model, \ie, $\frac{\partial v^{t+1}_{rj}}{\partial a^t_{rj}}=0$, $\frac{\partial a^{t+1}_{rj}}{\partial v^t_{rj}}=\psi^t_{rj}$, and $\frac{\partial a^{t+1}_{rj}}{\partial a^t_{rj}} = \rho - \psi^t_{rj}\beta$.

	        Using these values, we compute the new eligibility trace:
	        \begin{align}
	        \bm{\epsilon}^{t+1}_{rji} &= \frac{\partial\mathbf{h}^{t+1}_{rj}}{\partial\mathbf{h}^t_{rj}}\cdot\bm{\epsilon}^t_{rji} + \frac{\partial\mathbf{h}^{t+1}_{rj}}{\partial W_{rji}}\\
	        &= \begin{pmatrix}
	        \alpha\left(1 - z^t_{rj} - z_{rj}^{t-\delta t_\text{ref}}\right)\epsilon_{rji, v}^t + z_{ri}^{t-1}\\
	        \psi^t_{rj}\epsilon^t_{rji, v} + \left(\rho - \psi^t_{rj}\beta\right)\epsilon^t_{rji,a}
	        \end{pmatrix}\label{eq:ml_stdpalif_evector}
	        \end{align}
	        Note that the observable state of the afferent neuron $z_{ri}^{t-1}$ in Equation \ref{eq:ml_stdpalif_evector} changes to $z_{r-1, i}^t$ if $r > 1$ and $\bm{\epsilon}^{t+1}_{rji}$ corresponds to a weight between layer $r-1$ and $r$.
	        If the weight is instead between the network input and the first layer, then this $z_{ri}^{t-1}$ changes to $x^t_i$.
	        Note also that this corrects an inconsistency in the STDP-LIF model described in \citet{traub2020learning}, where the observable state at the current time step $t$ is accessed instead of $t-1$, thereby diverging from the e-prop model in Equation \ref{eq:ml_model}.

	        The eligibility trace remains
	        \begin{align}
	        e^{t}_{rji} &= \frac{\partial z^{t}_{rj}}{\partial\mathbf{h}^{t}_{rj}} \cdot \bm{\epsilon}^{t}_{rji}\\
	        &= \psi^t_{rj}\left(\epsilon^t_{rji, v} - \beta\epsilon^t_{rji, a}\right).
	        \end{align}

\begin{figure}[!ht]
	            \centering
	            \includegraphics[width=\linewidth]{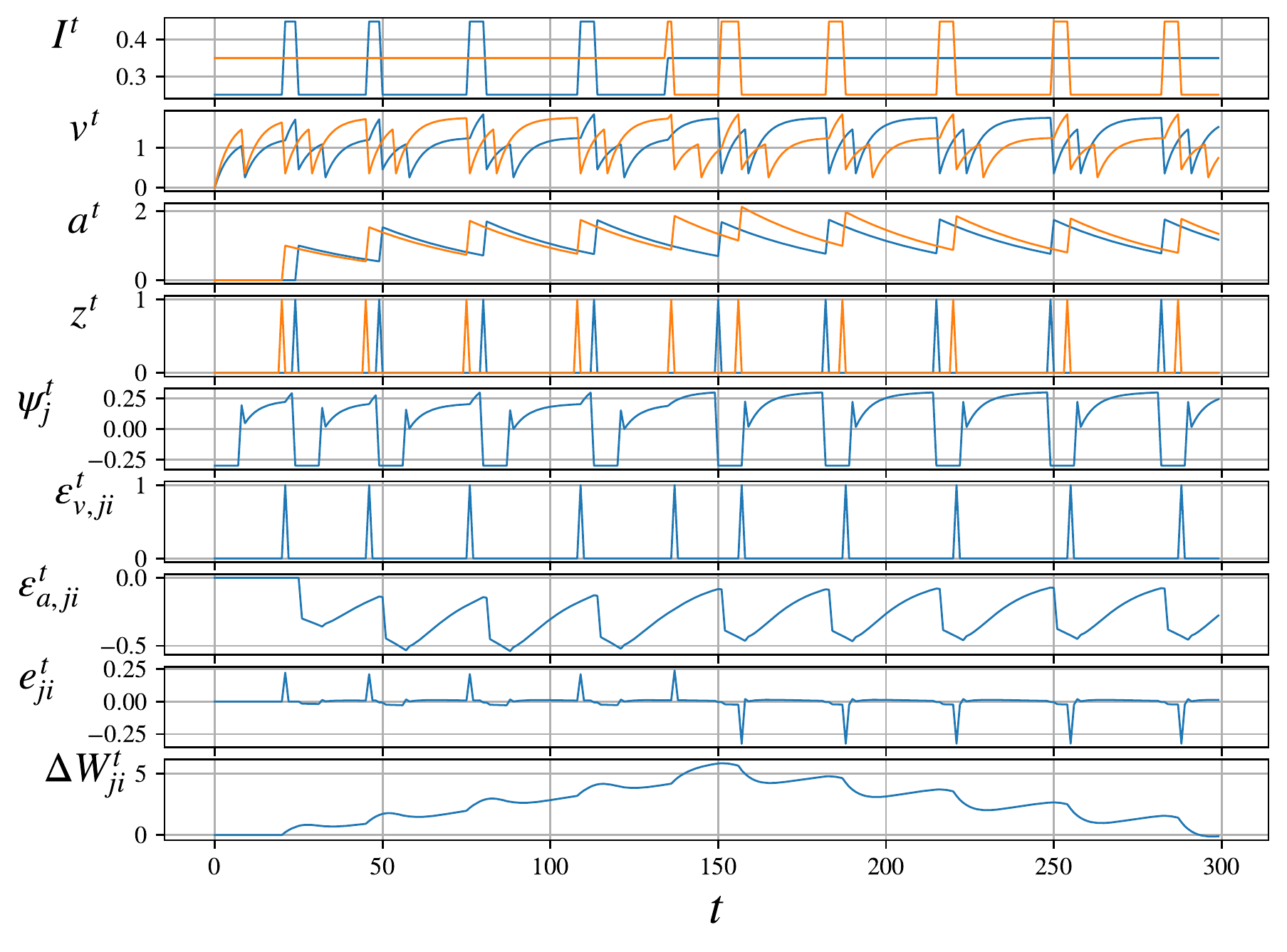}
	            \caption[Single-synapse STDP-ALIF simulation]{A single-synapse simulation of the STDP-ALIF neuron model. Orange and blue lines respectively describe properties of the afferent and efferent neuron.}
	            \label{fig:stdpalif}
	        \end{figure}

	        By resetting the neuron activation at spike time and after the refractory time, STDP is introduced in the system by clamping the pseudo-derivative to a negative value, instead of 0, during the refractory time:

	        \begin{equation}
	        \psi^t_{rj} = \begin{cases}
	        -\gamma&\mbox{if } t - t_{z_{rj}} < \delta t_\text{ref}\\
	        \gamma\max\left(0, 1 - \left|\frac{v^t_{rj}-v_\text{th}}{v_\text{th}}\right|\right)&\mbox{otherwise}
	        \end{cases}
	        \end{equation}
	        The factor of the pseudo-derivative and the eligibility vector can therefore produce both positive or negative eligibility traces and gradients (see Figure \ref{fig:stdpalif}).

		\subsubsection{Izhikevich neuron}\label{sec:izhikevich}

			The standard system of equations of the Izhikevich neuron is described by
			\begin{align}
			v' &= 0.04v^2 + 5v + 140 - a + I\\
			a' &= 0.004v - 0.02a,
			\end{align}
			where $v'$ and $a'$ are the values of the activation value $v$ and recovery variable $a$ at the next time point, and $I$ is the current input to the neuron.
			Following \citet{traub2020learning}, the activation reset and refractory period are built into this system of equations by replacing $v$ and $a$ by respectively:
			\begin{align}
			\tilde{v}^t_{rj} &= v^t_{rj} - \left(v^t_{rj} + 65\right)z^t_{rj}\\
			\tilde{a}^t_{rj} &= a^t_{rj} + 2z^t_{rj},
			\end{align}
			such that when a spike event occurs (\ie, $z^t_{rj} = 1$), the activation value is reset to its baseline value of $-65$, and the recovery variable increases by 2.
			We describe this ``self-resetting'' Izhikevich neuron in the context of multi-layer e-prop as follows:
			\begin{align}
			v^{t+1}_{rj} &= \tilde{v}^t_{rj} + 0.04\left(\tilde{v}^t_{rj}\right)^2 + 5\tilde{v}^t_{rj} + 140 - \tilde{a}^t_{rj} + I^t_{rj}\\
			a^{t+1}_{rj} &= \tilde{a}^t_{rj} + 0.004\tilde{v}^t_{rj}-0.02\tilde{a}^t_{rj}.
			\end{align}
			The partial derivatives of the hidden state $\mathbf{h}^t_{rj}$ can then be computed:
			\begin{align}
			\frac{\partial v^{t+1}_{rj}}{\partial v^t_{rj}} &= \left(1-z^t_{rj}\right)\left(6+0.08v^t_{rj}\right)\\
			\frac{\partial a^{t+1}_{rj}}{\partial v^t_{rj}} &= -1\\
			\frac{\partial v^{t+1}_{rj}}{\partial a^t_{rj}} &= 0.004\left(1-z^t_{rj}\right)\\
			\frac{\partial a^{t+1}_{rj}}{\partial a^t_{rj}} &= 0.98.
			\end{align}

			Using these values, we compute the new eligibility trace:
	        \begin{align}
	        \bm{\epsilon}^{t+1}_{rji} &= \frac{\partial\mathbf{h}^{t+1}_{rj}}{\partial\mathbf{h}^t_{rj}}\cdot\bm{\epsilon}^t_{rji} + \frac{\partial\mathbf{h}^{t+1}_{rj}}{\partial W_{rji}}\\
	        &= \begin{pmatrix}\left(1-z^t_{rj}\right)\left(6+0.08v^t_{rj}\right)\epsilon^t_{rji, v} -\epsilon^t_{rji, a} + z_{ri}^{t-1}\\
	        0.004\left(1-z^t_{rj}\right)\epsilon^t_{rji, v} + 0.98\epsilon^t_{rji, a}
	        \end{pmatrix}\label{eq:ml_izhikevich_evector}
	        \end{align}
	        As in \citet{traub2020learning}, the pseudo-derivative is defined as
	        \begin{equation}
	        \psi^t_{rj} = \gamma\ \exp\left(\frac{\min\left(v^t_{rj}, 30\right) - 30}{30}\right),
	        \end{equation}
	        such that
	        \begin{equation}
	        \begin{pmatrix}\frac{\partial z^t_{rj}}{\partial v^t_{rj}}\\\frac{\partial z^t_{rj}}{\partial u^t_{rj}}\end{pmatrix}
	        = \begin{pmatrix}\psi^t_{rj}\\0\end{pmatrix},
	        \end{equation}
	        and therefore only $\epsilon^t_{rji, v}$ is used in computing the eligibility trace:
	        \begin{align}
	        e^t_{rji} &= \left(\psi^t_{rj}\ 0\right)\begin{pmatrix}\epsilon^t_{rji, v}\\\epsilon^t_{rji, a}\end{pmatrix} \\
	        &= \psi^t_{rj}\epsilon^t_{rji, v}.
	        \end{align}

\begin{figure}[!ht]
		    \centering
		    \includegraphics[width=\linewidth]{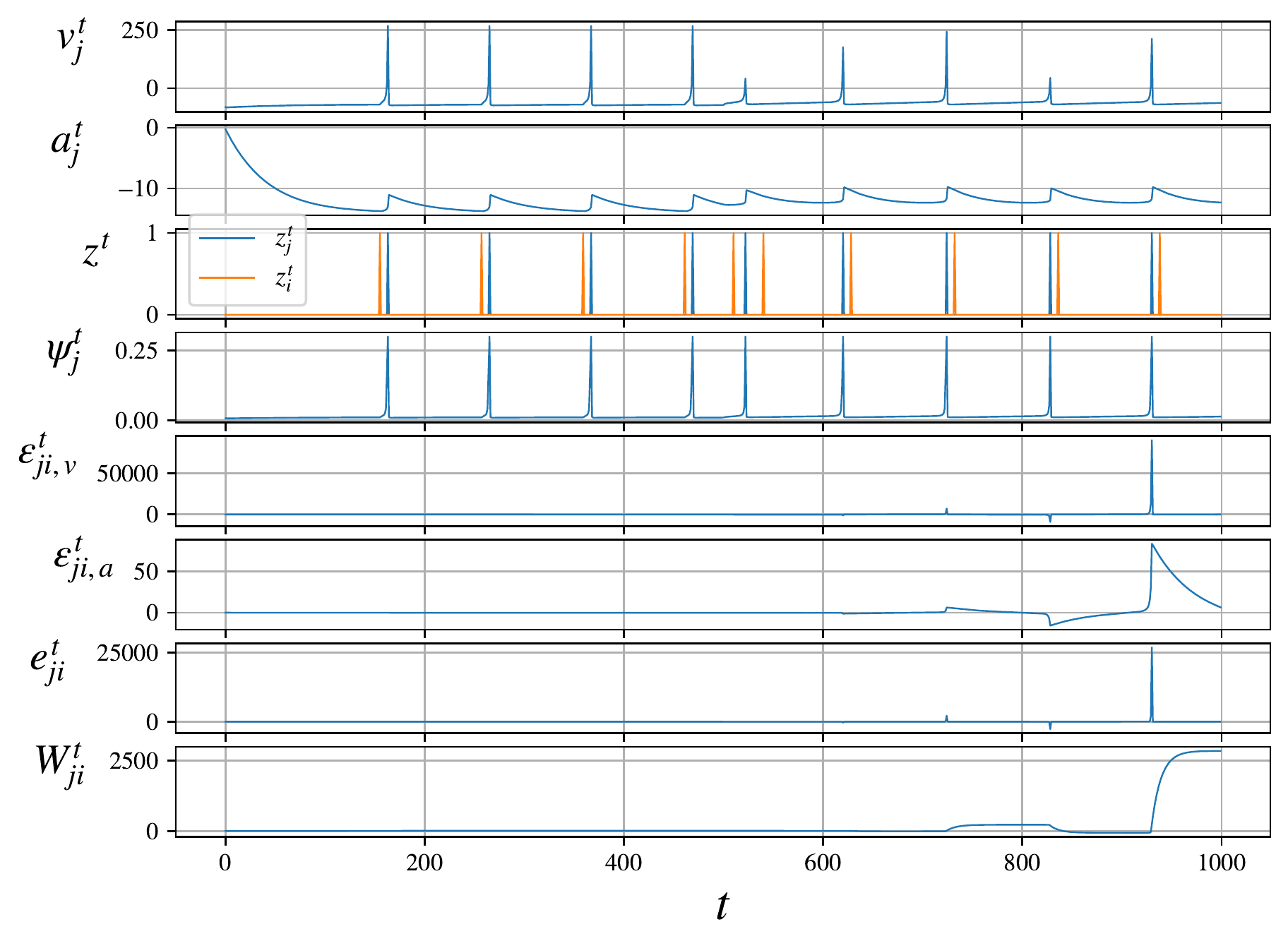}
		    \caption[Uncorrected Izhikevich neuron simulation]{A single-synapse simulation of the uncorrected Izhikevich neuron. Note that $\epsilon^t_{ji, v}$ takes on extreme values and that the eligibility vector flips sign at any pair of spikes. Orange and blue lines respectively describe properties of the afferent and efferent neuron.}
		    \label{fig:demo_izh}
		\end{figure}

	    However, when inserting these equations in a single-synapse demo, the eligibility vector assumes extremely high or low values (see Figure \ref{fig:demo_izh}).

\begin{figure}[!ht]
            \centering
            \includegraphics[width=\linewidth]{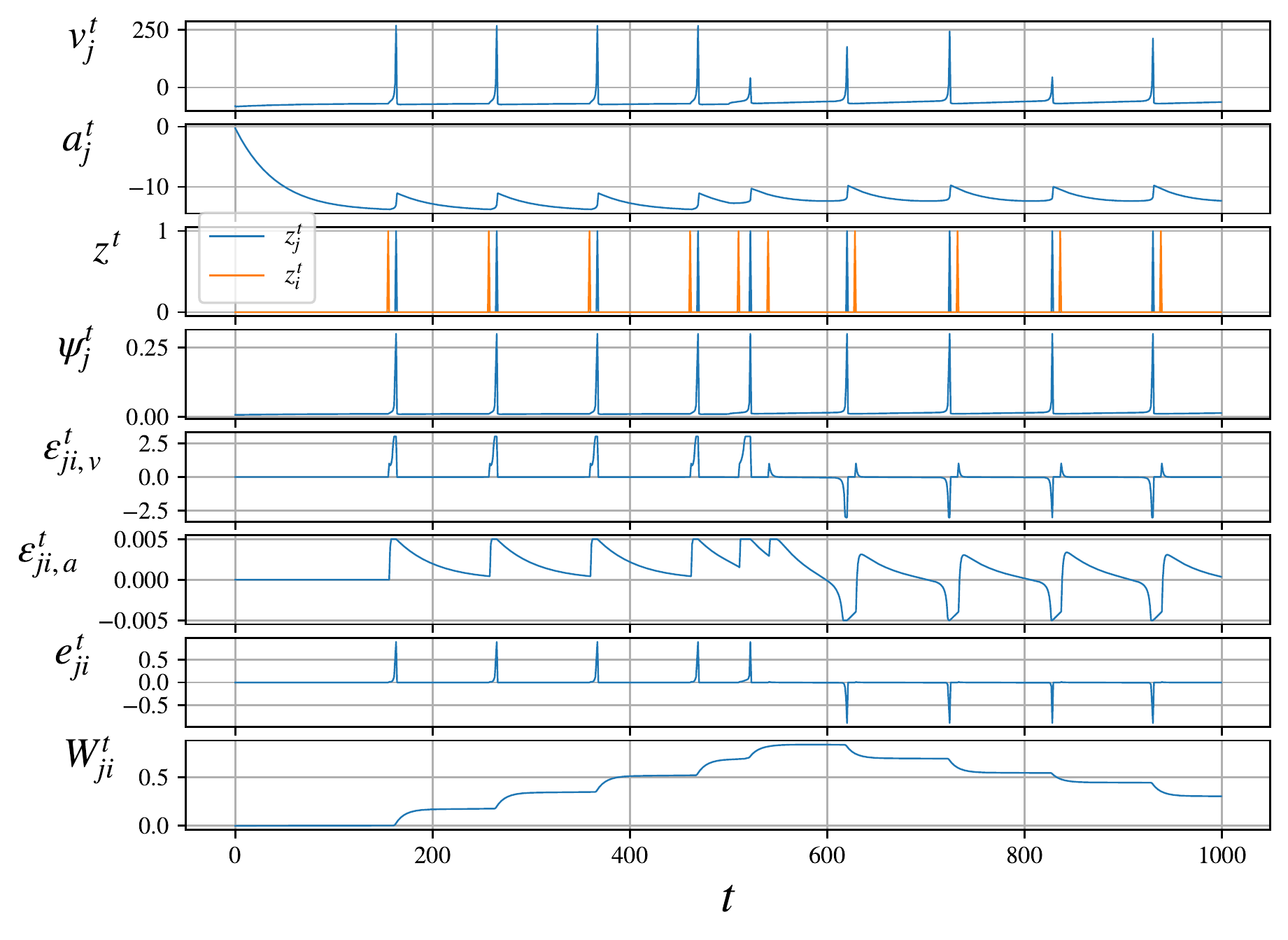}
            \caption[Corrected Izhikevich neuron simulation]{A single-synapse simulation of the corrected Izhikevich neuron. Orange and blue lines respectively describe properties of the afferent and efferent neuron.}
            \label{fig:demo_izh_corrected}
        \end{figure}

	    This suggests that the Izhikevich neuron does not fit the e-prop framework well.
	    In this report, this is corrected by clipping the value of $\epsilon^t_{ji,v}$ to $\left[-3, 3\right]$ and $\epsilon^t_{ji,a}$ to $\left[-0.005, 0.005\right]$.
	    This correction yields the desired STDP behavior (see Figure \ref{fig:demo_izh_corrected}).

\newpage
\section{Regularization}

	Firing rate regularization and L2 regularization are applied to improve the stability of the learning process and the generalizability of the resulting model.
	These two regularization methods are motivated by biological plausibility, ease of implementation in the e-prop framework, and improved empirical performance.

	\subsection{Firing rate regularization}
		A firing rate regularization term is added to individually modulate the spike frequencies of the neurons.
		Since spikes in a neuromorphic architecture cost energy, the practical motivation for this regularization term is that it increases the energy efficiency of the model.
		The biological motivation is that the firing rate of biological neurons is also under homeostatic control \citep{erickson2006activity}.

		Firing rate is implemented by adding a regularization term $E_\text{reg}$ to the loss function that penalizes neurons that have a too low or too high firing rate:
		\begin{equation}
			E_\text{reg} = \frac{1}{2}\sum_j\left(f^\text{target} - f^{\text{av}, t}_{rj}\right)^2,
		\end{equation}
		where $f^\text{target}$ is a target firing rate of \SI{10}{\Hz}, and
		\begin{equation}
		f^{\text{av},t}_{rj} = \frac{1}{t} z^{\text{total},t}_{rj}
		\end{equation}
		is the running average of the spike frequency, where $z^{\text{total},t}$ accumulates spikes emitted by neuron $j$ in layer $r$ up to (and including) time step $t$.
		Note that $z^{\text{total},0} = 0$, \ie, the accumulation resets at each new training sample.
		By implementing this sum as a hidden variable, e-prop remains an online and local training algorithm when firing rate regularization is implemented.
		Adhering to these two constraints supports the biological plausibility of the firing rate regularization term.
		After the training sample, the effect of the firing rate regularization on the weight update is integrated.

		Another possibility to compute the average firing rate would be to track an exponentially decaying firing rate.
		This was not implemented in this research for the following two reasons. First, the effects of the firing rate regularization are integrated only at the end of a training sample (see Equation \ref{eq:deltareg}), likely causing any fluctuations of the average frequency over time to cancel each other out and result in an effectively similar regularization term as the current, non-decaying frequency calculation.
		Second, since one of the objectives of this research is to reproduce results obtained in \citet{bellec2020solution} and expand on that paper, the regularization term is kept as faithful to theirs as possible.
		However, analyzing the effects of different firing rate regularization terms might be an interesting direction for future research.

		To insert the regularization term into the e-prop framework, we compute the weight update that regularizes the firing rate toward $f^\text{target}$ through gradient descent, similarly to the main e-prop weight update in Equation \ref{eq:eprop_grd}:
		\begin{equation}
		\frac{\partial E_\text{reg}}{\partial z_{rj}^t} = \left(f^\text{target} - f^{\text{av}, t}_{rj}\right).
		\end{equation}
		Note that this regularization loss differs from the firing rate regularization described in \citet{bellec2020solution}, in which the firing rate is calculated in an offline fashion, by retroactively computing the average firing rate based on all spikes instead of only accumulated spikes.
		Note also that in \citet{bellec2020solution}, $\frac{\partial E_\text{reg}}{\partial z_{rj}^t}$ is multiplied with the eligibility trace $e^t_{rji}$, as in Equation \ref{eq:eprop_grd} to obtain the weight update, whereas in this report, the eligibility trace is omitted, resulting in a number of benefits:
		\begin{enumerate}
			\item It allows silent neurons that have infrequently spiking afferent neurons to more easily increase their firing rate, because their low afferent eligibility traces no longer nullify the regularization gradients, and thereby result in a better empirical learning performance.
			\item It is more efficient in emulations on von Neumann machines, because the element-wise multiplication of $\frac{\partial E_\text{reg}}{\partial z_{rj}^t}$ and the eligibility trace is a relatively large computation on the order $\Theta\!\left(n^2\right)$ that no longer needs to be computed.
			\item It is more intuitive, as only the gradient of the firing rate is used to compute the weight update.
		\end{enumerate}
		We apply the weight update $\Delta W_{rji}$ of the regularization gradient using
		\begin{equation}\label{eq:deltareg}
		\Delta_\text{reg} W_{rji} = -\eta\ c_\text{reg}\sum_t\left(f^\text{target} - f^{\text{av}, t}_{rj}\right).
		\end{equation}
		Note that the regularization gradients can be combined and accumulated over time on the same synaptic variable as the normal gradients, facilitating practical implementation of the learning procedure in both software emulations and neuromorphic embeddings:
		\begin{equation}
		\Delta W_{rji} = -\eta\ \sum_t\left(c_\text{reg}\left(f^\text{target} - f^{\text{av}, t}_{rj}\right) + L^t_{rj}\cdot\bar{e}^t_{rji}\right).
		\end{equation}

	\subsection{L2 regularization}
		To regularize weights around 0, a small fraction (parametrized by $c_\text{L2}$) of the value of the weight is added to its gradient value at every weight update:
		\begin{equation}
		\Delta_\text{L2} W_{rji} = -\eta\ c_\text{L2} \cdot W_{rji},
		\end{equation}
		which can be added to the full weight update as an extra term:
		\begin{equation}
		\Delta W_{rji} = -\eta\left(c_\text{L2} \cdot W_{rji} + \sum_t\left(c_\text{reg}\left(f^\text{target} - f^{\text{av}, t}_{rj}\right) + L^t_{rj}\cdot\bar{e}^t_{rji}\right)\right).
		\end{equation}
		The statistical motivation for this extra regularization term is that by softly restricting the weights, the network is less likely to overfit on the training data.

		The biological motivation is that biological synapses are regularized by a multiplicative factor to decrease the strength of individual synapses to the same proportion \citep{turrigiano2000hebb}, likely counteracting the run-away effects that the positive feedback of STDP naturally induces \citep{siddoway2014molecular}.
		Moreover, there are natural bounds of the strength of a biological synapse, measured by the amplitude of the postsynaptic potential, because the number of neurotransmitter vesicles and release sites is physically limited \citep{del1954quantal}.
		These natural limits are approximately \SI{0.4}{\mV} to \SI{20}{\mV} \citep{diaz2006target}.



\section{Optimizer}\label{sec:adam}

	For simplicity, in this report weight updates are described using stochastic gradient descent:
	\begin{equation}
		\Delta W_{rji} = -\eta \sum_t L^t_{rj}\cdot\bar{e}^t_{rji}.
	\end{equation}
	However, the results described in Section \ref{sec:results} are obtained using Adam (or Adaptive Moment Estimation) \citep{kingma2014adam}.
	This optimization method tracks running averages of the gradient and its second moment (resp. $M_{rji}$ and $V_{rji}$), and fits the local and online constraints of e-prop, because the running averages are tracked per individual synapse.
	The Adam weight update in the context of multi-layer e-prop is given by:
	\begin{align}
	M_{rji}^{(i+1)} &= \beta_1 M_{rji}^{(i)} + \left(1 - \beta_1\right)G^{(i)}_{rji} \\
	V_{rji}^{(i+1)} &= \beta_2 V_{rji}^{(i)} + \left(1 - \beta_2\right)\left(G^{(i)}_{rji}\right)^2 \\
	\widehat{M}_{rji} &= \frac{M_{rji}^{(i+1)}}{1 - \beta_1^{i+1}} \\
	\widehat{V}_{rji} &= \frac{V_{rji}^{(i+1)}}{1 - \beta_2^{i+1}} \\
	\Delta W_{rji}^{(i+1)} &= -\eta \frac{\widehat{M}_{rji}}{\sqrt{\widehat{V}_{rji}} + 10^{-5}},
	\end{align}
	where $G^{(i)}$ is the estimated gradient $\sum_t L^t_{rj}\cdot\bar{e}^t_{rji}$ at weight update $i$, and $\beta_1=0.9$ and $\beta_2=0.999$ are the forgetting factors for the gradient and its second moment, respectively.
	The firing rate and L2 regularization terms are omitted here for clarity.
	Note that the forgetting factors are not indexed, but raised to the power of $i+1$, in computing the bias-corrected estimates $\widehat{M}_{rji}$ and $\widehat{V}_{rji}$.
	Note also that minibatches of size 32 are used to more accurately estimate the gradient and enable a stabler descent in the error landscape.
	The value of $G^{(i)}_{rji}$ is computed as the mean over the minibatch.

	Note that the learning rate is linearly ramped up from 0 to $\eta$ during the first epoch, such that the initial minibatches are used to aggregate good initial momentum buffers, as the variance is higher when fewer minibatches are processed.
	This ``warming up'' of the learning rate is a variance reduction technique that has shown beneficial results in training other models \citep{liu2019variance}.
	Empirical observations on the resulting learning curves (see Section \ref{sec:results}) suggest that this procedure does not hamper a rapid initial decrease of the loss function.

%% file: Chapters/Chapter04.tex
\chapter{Discussion}\label{ch:discussion}
In this chapter, the learning performance and regularization behavior of the ALIF, STDP-ALIF, and Izhikevich neurons are compared and discussed.
Then, the effect of stacking multiple recurrent layers on the learning performance and speed is examined.
Next, possible future research avenues are discussed.
\section{Results}\label{sec:results}

\begin{figure}[ht]
	    \myfloatalign
	    \includegraphics[width=\linewidth]{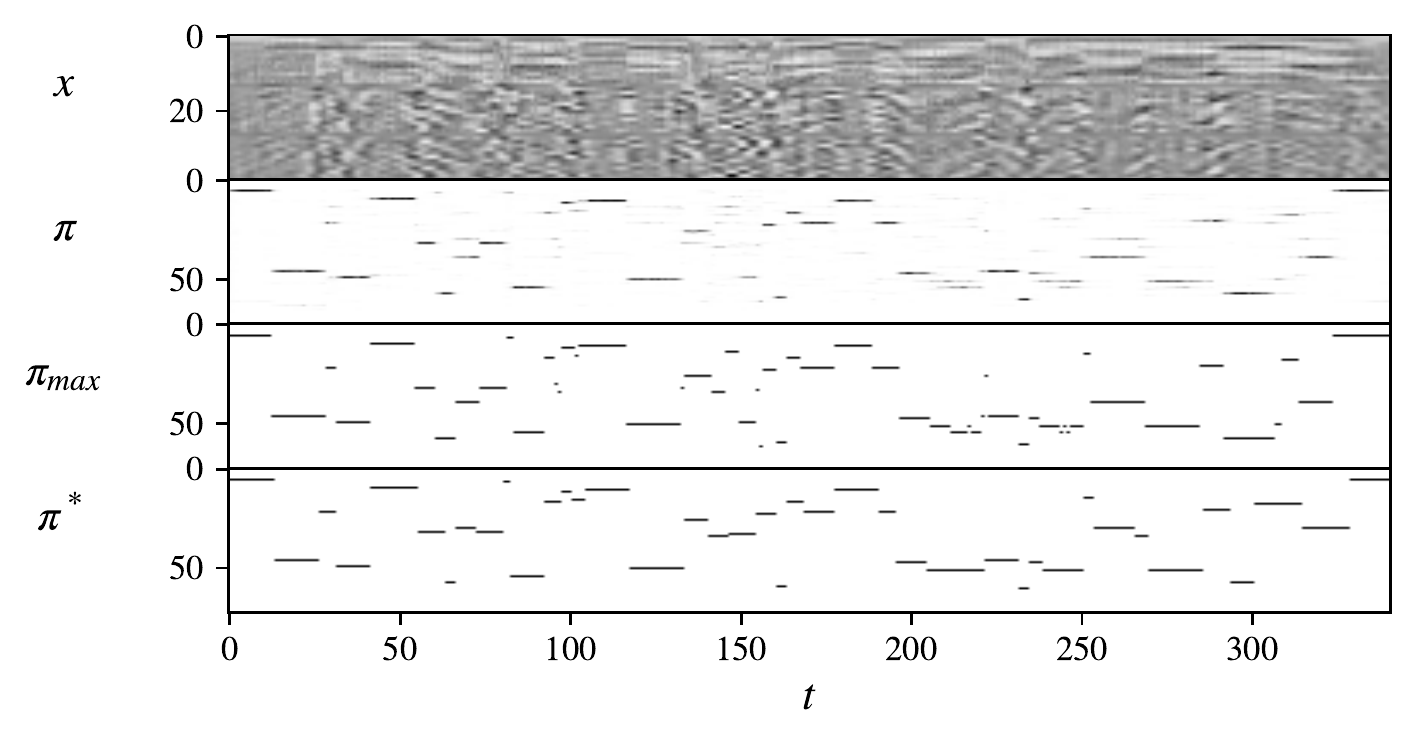}
	    \caption[Input/output/target example]{An example validation result using a trained ALIF model. The plot in the top row shows the standardized MFCC frames including its first and second derivatives of a sentence changing over time. The plot in the second row shows the probability distributions $\pi$ of the frame-wise outputs of the model. The plot in the third row indicates the predicted phone $\pi_\text{max}$ per frame. The plot in the last row shows the target phones $\pi^*$.}
	    \label{fig:inoutpair}
	  \end{figure}

	Figure \ref{fig:inoutpair} shows a typical classification result of a full validation sentence.

	\subsection{Comparing neuron models}

\begin{figure}[bth]
			    \myfloatalign
			    \subfloat[Percentage of samples wrongly classified.]
			    {\label{fig:percwrong}\includegraphics[height=5cm, keepaspectratio]{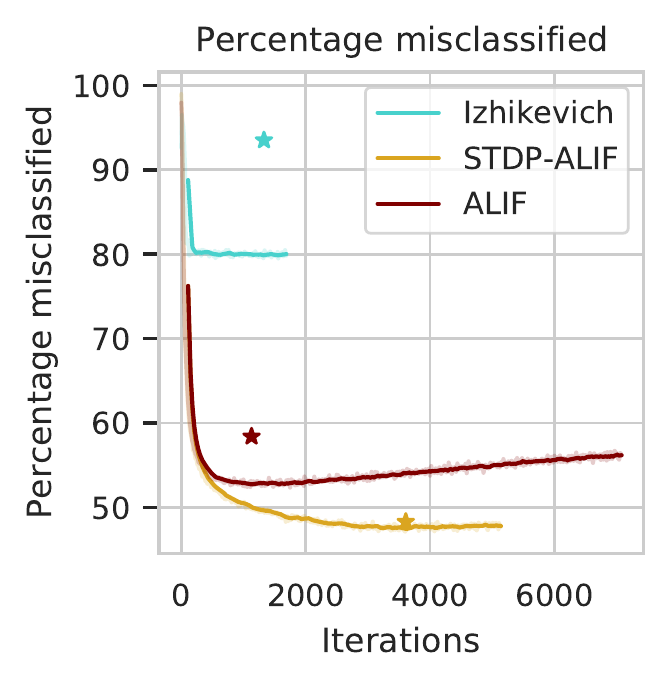}} \quad
			    \subfloat[Cross-entropy loss (log-scaled).]
			    {\label{fig:crossentropy}%
			        \includegraphics[height=5cm, keepaspectratio]{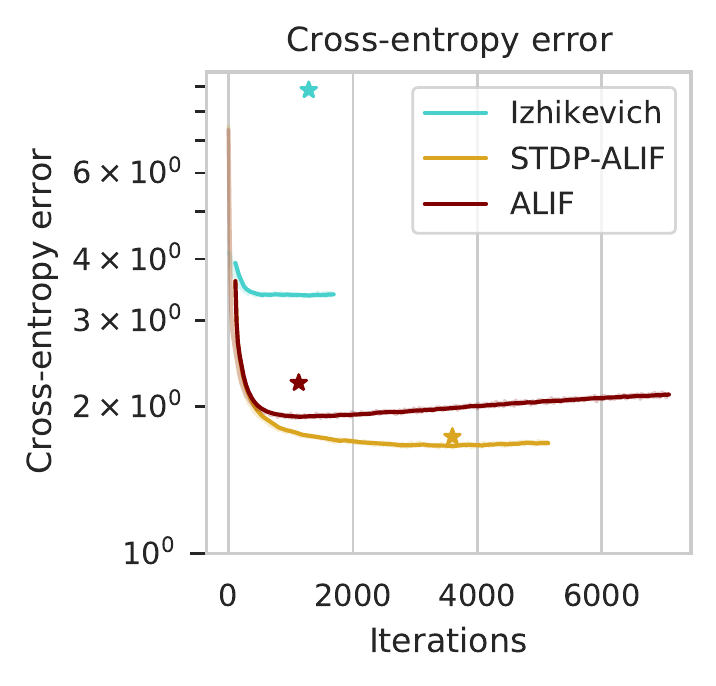}}
			    \caption[Single-layer classification performance per neuron model]{Classification performance on the validation data for each of the three neuron models in a single-layer e-prop model. The opaque lines indicate the running average of the real validation scores indicated by the transparent lines. The star symbols indicate the performances on the test set, with a misclassification rate of 93.5\% for the Izhikevich neuron, 58.4\% for the ALIF neuron, and 48.3\% for the STDP-ALIF neuron type.}\label{fig:sl-acc}
			\end{figure}

		\paragraph{Accuracy}
			The main outcome of the neuron model comparison is that in these results, the STDP-ALIF neuron outperforms the ALIF and corrected Izhikevich neuron models in classifying phones in the TIMIT dataset.
			This suggests that including STDP-like behavior to the ALIF neuron results in a better learning performance, answering the primary research objective posed in Chapter \ref{ch:introduction}.
			Furthermore, the ALIF and STDP-ALIF neuron perform better than the Izhikevich e-prop neuron model.
			In Figure \ref{fig:percwrong} the Izhikevich neuron reaches a misclassification rate of 93.8\% on the test set, which is only slightly better than constantly predicting the most frequent class.
			The ALIF neuron model reaches a test misclassification rate of 58.4\% in relatively few iterations, after which validation performance starts to decrease.
			The STDP-ALIF neuron model scores best, reaching a performance of 48.3\% after approximately 3500 iterations, suggesting that the addition of the STDP mechanism to the ALIF neuron improves the classification performance.
			Furthermore, the STDP-ALIF model does not show signs of overfitting as much as the ALIF neuron such as a decreasing validation performance in Figure \ref{fig:percwrong}.
			However, the Izhikevich neuron shows STDP behavior too but performs poorly, suggesting that STDP by itself does not necessarily constitute a well-performing neuron model.
			The STDP-ALIF neuron may instead work by virtue of another factor, such as its better spike frequency adaptation compared to the Izhikevich neuron model.
			Note that the test performance was obtained from the model with the best validation accuracy (the used hyperparameters are listed in Table \ref{tab:hparams}).

			Figure \ref{fig:crossentropy} illustrates the decrease of the cross-entropy score, which for the ALIF and STDP-ALIF neurons is comparable to that of the misclassification rate.
			The cross-entropy and classification performance of the Izhikevich neuron stalls relatively quickly at poor levels, suggesting that it trains its bias toward more frequent phone classes in the training data rather than learning a general relationship between input MFCCs and classes.

\begin{figure}[bth]
			    \myfloatalign
			    \subfloat[Mean spiking frequency. Note the logarithmic horizontal axis.]
			    {\label{fig:freqs}\includegraphics[height=5cm, keepaspectratio]{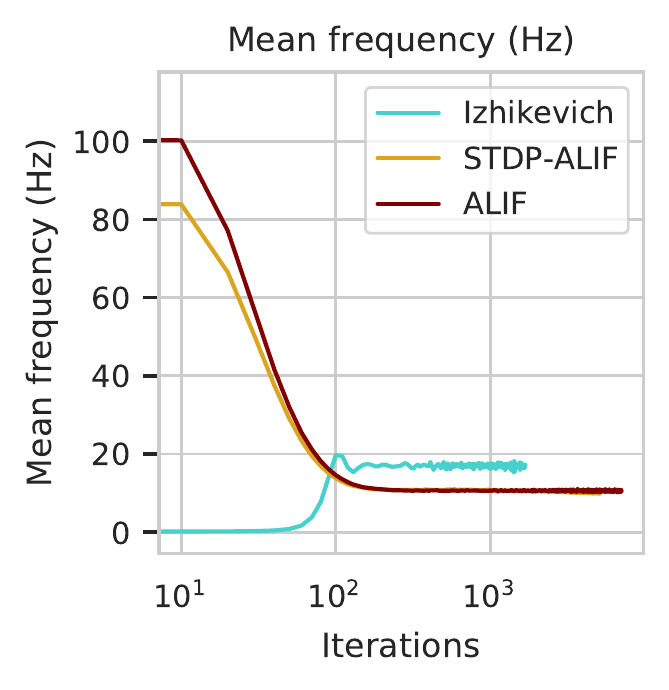}} \quad
			    \subfloat[Regularization error. Note the logarithmic vertical axis.]
			    {\label{fig:regerr}%
			        \includegraphics[height=5cm, keepaspectratio]{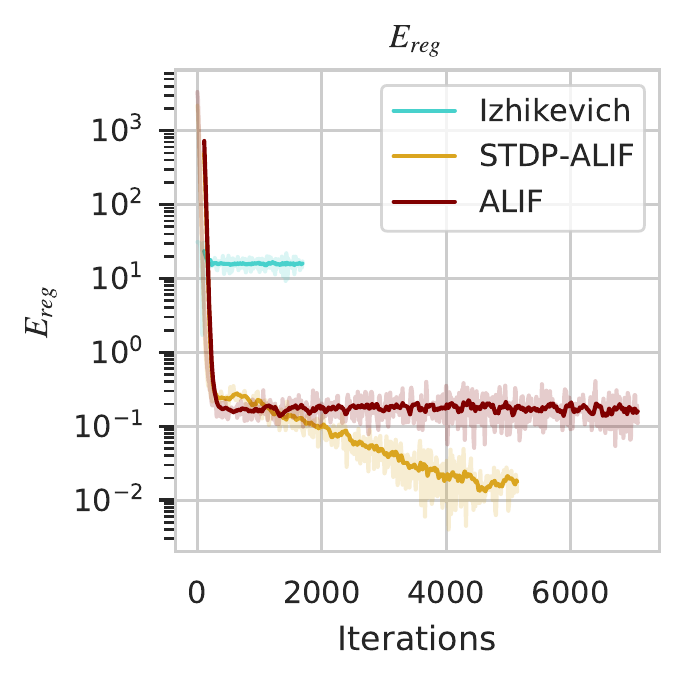}}
			    \caption[Single-layer firing rate regularization per neuron model]{Effect of firing rate regularization on the validation data for each of the three neuron models.}\label{fig:sl-reg}
			\end{figure}

		\paragraph{Firing rate}
			Figure \ref{fig:freqs} illustrates the effect of the firing regularization term.
			It can be observed that the ALIF and STDP-ALIF neuron models are able to quickly modulate their mean spiking frequencies to the desired target frequency of \SI{10}{\Hz}, but the Izhikevich neuron overshoots to a mean spiking frequency of approximately \SI{18}{\Hz}.

			Figure \ref{fig:regerr} illustrates the decrease of the regularization error.
			The regularization error of the Izhikevich and ALIF neuron models quickly converges to fluctuate around a constant value, whereas that of the STDP-ALIF neuron model continues to decrease over time, even after the mean spiking frequency and classification performance have both converged to a plateau.

	\subsection{Comparing network depth}
\begin{figure}[bth]
		    \myfloatalign
		    \subfloat[ALIF model.]
		    {\label{fig:ml-pwrong-alif}\includegraphics[height=5cm, keepaspectratio]{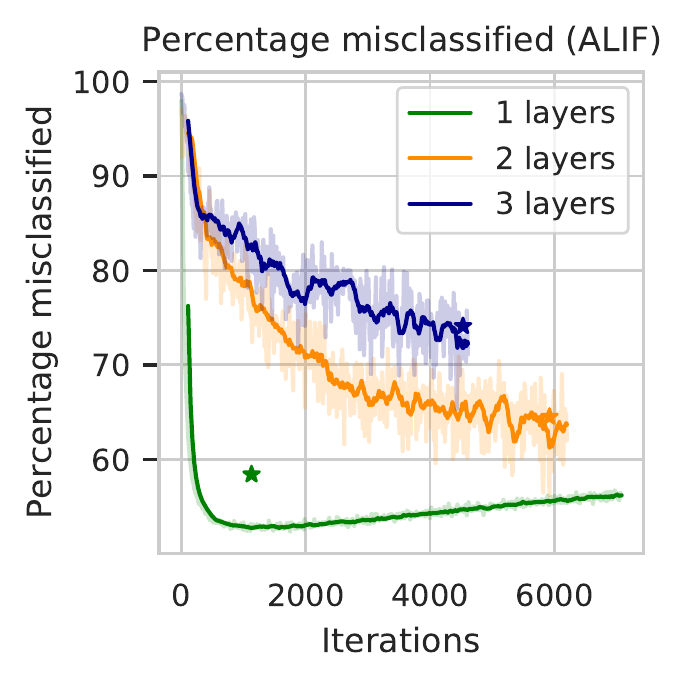}} \quad
		    \subfloat[STDP-ALIF model.]
		    {\label{fig:ml-pwrong-stdpalif}%
		        \includegraphics[height=5cm, keepaspectratio]{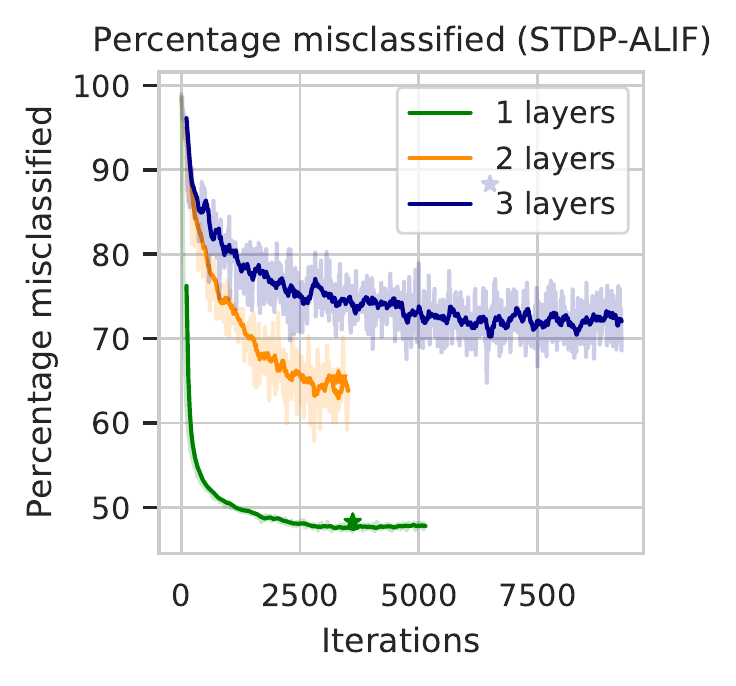}} \\
		    \subfloat[Izhikevich model.]
		    {\label{fig:ml-pwrong-izh}\includegraphics[height=5cm, keepaspectratio]{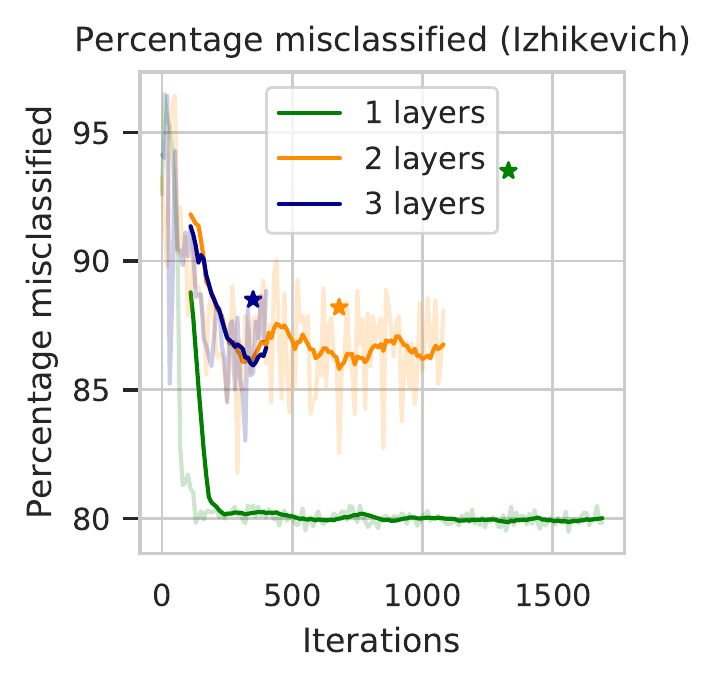}}
		    \caption[Single- and multi-layer accuracy comparison]{Accuracy comparison on the validation data between single- and multi-layer e-prop models.}\label{fig:ml-percwrong}
		\end{figure}

		The comparison between the network depth in Figures \ref{fig:ml-pwrong-alif}--\ref{fig:ml-pwrong-izh} suggests that single-layer e-prop networks train considerably more efficiently and accurately than multi-layer e-prop networks and show less variance among validation runs.
		This holds for all tested neuron types.
		The cross-entropy error, spiking frequency, and regularization error are also better for single-layer networks (see Figure \ref{fig:ml-otherresults}).

		Therefore, rearranging the neurons into a stacked architecture does not appear to improve the classification performance, answering the secondary research objective posed in Chapter \ref{ch:introduction}.
		In particular, it appears to diminishes the learning speed to a significant extent and render multi-layer e-prop architectures more inefficient than single-layer ones.
		However, it is not certain that multi-layer architectures are necessarily worse---they train more slowly, but in this report the performance was still improving when their training runs were interrupted due to practical limits.
		Therefore, particularly for the ALIF neuron, the multi-layer networks might outperform the single-layer networks if future work where low computing power and energy costs are not a priority.

\section{Possible improvements}
    There are many hypothetical ways of improving the performance or biological plausibility of e-prop that have not yet been considered in this report.
    For instance, a likely reason that the learning speed of the multi-layer architectures was slower than their single-layer counterparts is that the weights are poorly initialized.
    Empirical observation of the learning process suggested that during early epochs, spiking activity faded in deeper layers, because the spiking activity from a preceding layer is generally weaker than the input values the first layer receives.
    Higher weights in-between layers mitigate this fading activity, but require some search to find a good value.
    In this report, the firing rate regularization term approximated this value, but learning is more efficient with a better initialization, since initial synaptic weights significantly affect the performance of STDP-based SNNs \citep{kim2020initial}.

    Also in this report, certain parameters such as firing rate targets ($f^\text{target}$), activity leak ($\alpha$), and feedback signals were constant for all neurons, except the threshold adaptivity ($\beta$), which was 0 for a randomly selected 25\% of the neurons to emulate non-adaptive LIF neurons.
    Future research could examine the effects of sampling some of these parameters from a distribution for each neuron, thereby creating a more diverse population of neurons with different time scales.
    This sampling requires a careful assessment of the time scales related to the learning task; in particular, the network should be able to process the slowest relevant time scale of the task \citep{jaeger2021dimensions}.
    In the TIMIT classification task, for instance, this could span the whole time fragment, because initial words give semantic context to all subsequent words.
    However, a more predictive time scale for the TIMIT dataset is on the scale of approximately 8 MFCC frames, which is now accurately captured by the adaptive threshold component in the ALIF neuron model.
    Temporal dependencies can also be found when moving in the opposite direction---later words and sounds give informative context to earlier words and sounds.
    In \citet{bellec2020solution}, this context was captured in a bidirectional network, improving the accuracy by nearly 15\%.
    In this report, this was empirically validated as well, but left undiscussed because a bidirectional network is not biologically plausible, as directly accessing future input values violates the ``online'' constraint.
    According to \citet{bellec2020solution}, e-prop suggests that the experimentally found diverse time constants of the firing activity of populations of neurons in different brain areas \citep{runyan2017distinct} are correlated with their capability to handle corresponding ranges of delays in temporal credit assignment for learning.
    Setting different values for these parameters per layer might also have a beneficial effect; \citet{ahmed1998estimates} suggested that deeper layers display slower and weaker adaptation rates than early layers.

    In the brain, neurons primarily tend to connect to nearby neurons.
    This suggests that the effects of the topology within a layer might positively affect the learning process.
    A simple lattice topology might better approximate the connectivity of the brain, decrease the computational complexity in emulations in von Neumann machines, and allow easier on-chip implementations in neuromorphic hardware.
    Hierarchical clustering of neurons might also have a beneficial effect, as this has been demonstrated to improve R-STDP in SNNs \citep{weidel2021unsupervised} and address the scalability issue of SNNs \citep{carrillo2012scalable}.
    Because a neuromorphic system can support complex network operations \citep{hasler1990vlsi}, large-scale conductance-based SNNs \citep{yang2019scalable,Yang2019RealTimeNS} and asynchronous communication in VLSIs through address-event-representation \citep{lazzaro1993silicon,deiss1999pulse} might be suitable to further customize the connectivity graph of an e-prop architecture.

    Other exciting research avenues include connectivity graphs that change over time through a dynamic pruning and growing of weights and neurons.
    Here, the biological motivation is that the human brain prunes synaptic connections during early development \citep{huttenlocher1979synaptic}.
    \citet{elbez2020progressive} demonstrated that 75\% of a SNN can be compressed while preserving its performance, but it is not clear if this can be applied in a biologically plausible way in the e-prop framework.
    However, integrating stochastic synaptic rewiring \citep{kappel2018dynamic} into an ALIF network can improve its short-term memory \citep{bellec2020solution}.

    Finally, synaptic delay might improve the temporal processing power of an e-prop model.
    In this report, communication between neurons was transmitted as a spike over a synapse with a delay of 1 ms.
    This delay could differ among synapses, such that potentially informative past inputs are more accurately preserved in synaptic delays, rather than only in eligibility traces and activity loops.
    This can help deal with tasks that require processing information on multiple time spans \citep{jaeger2021dimensions}.
    This resembles the variable physical length of myelinated biological synapses and the number of nodes of Ranvier along them, affecting the conductance of the action potential \citep{bean2007action}.

\section{Future directions}
    As neuromorphic computing matures, neuroscience improves, and DL increasingly hits fundamental limitations, there is an exciting future for biologically plausible SNNs.
    There is much to gain from cross-fertilization between these fields.
    The popularity of DL was accelerated by accessible platforms to implement and deploy ANNs.
    Similar high-level simulation platforms are now in active development, which can integrate the typical behavior of memristive device models into crossbar architectures within DL systems \citep{lammie2020memtorch}.

    Recent advances in neuromorphic computing indicate this increasing popularity.
    Neuromorphic architectures have been used for mapless navigation with 75 times lower power and better performance \citep{tang2020reinforcement}; as low-power solutions for simultaneous localization and mapping of mobile robots \citep{tang2019spiking}, for planning \citep{fischl2017path}, and control \citep{blum2017neuromorphic}; and self-repairing SNN for fault detection \citep{zhu2017target}.
    While cross-fertilization between neuromorphic computing and quantum computing is starting to take place \citep{russek2016stochastic}, as quantum superposition and entanglement can be used to process information in parallel and in a high-dimensional state space \citep{fujii2017harnessing, yamamoto2017coherent,tacchino2019artificial}, more physics and materials science is required to build efficient neuromorphic architectures \citep{markovic2020physics}.
    The same holds for the cross-fertilization between neuroscience and learning rules of biologically plausible SNNs.
    Nanodevices that emulate biological synapses with learning functions can benefit neuromorphic architectures \citep{yao2017face,wang2018photonic,ren2018analytical}, particularly the two-terminal memristor \citep{jo2010nanoscale,wang2017memristors}.
    However, it has been argued that the learning principles of biological NNs are not explored enough to design engineering solutions \citep{gorban2019unreasonable,taherkhani2018supervised}.
    Feedback connections, for which the brain uses neurotransmitters, may become particularly problematic in large-scale neuromorphic systems.
    Another issue in analog computation is how to match the system's internal temporal processing to that of its inputs.
    Emulating neural dynamics on a physical substrate is more efficient but requires constraints to match the brain's timescales \citep{mead1990neuromorphic,jaeger2021dimensions}.
    Future work on e-prop could explore a combination with attention-based models in order to cover multiple timescales \citep{bellec2020solution}.

%% file: Chapters/Chapter06.tex
\chapter{Conclusion}\label{ch:conclusion}

As deep learning models require increasing amount of energy, the upcoming neuromorphic computing paradigm offers various ways to more efficiently run spiking neural networks.
Spiking neural networks can learn to perform tasks with a good performance and low energy requirements, but there is no established learning algorithm yet.
In this paper, the e-prop learning algorithm for recurrent spiking neural networks was combined with STDP, which is a major component of biological learning.

In this report, the e-prop framework was applied on the TIMIT phone classification task, meeting the objectives listed in Chapter \ref{ch:introduction}.
First, the performance of the ALIF neuron was reproduced using the explicit e-prop equations.
Next, the STDP-LIF neuron was modified to the STDP-ALIF model that was experimentally verified to outperform the ALIF neuron on the TIMIT learning task.
Also, the Izhikevich neuron, which also shows STDP behavior, was shown to be unstable and performing worse than the ALIF and STDP-ALIF neurons.
This suggests that STDP does not provide an adequate neuron model by itself, but that e.g. spike frequency adaptation also needs to be taken into account.
However, enhancing an already well-performing neuron model to display STDP-like properties can improve the performance.

Finally, the effect of stacking multiple layers was also examined in combination with the ALIF, STDP-ALIF, and Izhikevich neuron model, and did not appear to improve the learning performance in this task.

Possible future work on this topic includes research on the effects of more elaborate weight initialization methods, variable hyperparameters for individual neurons, different static or dynamic connectivity graphs, and synaptic delays.

The scientific gain of this research is that the link between STDP and e-prop was more closely examined than in previous literature, and that the inclusion of STDP in e-prop can lead to a more accurate or efficient learning performance.
E-prop combined with the STDP-ALIF neuron model remains a framework that offers high potential for biologically plausible learning algorithms for SNNs, which can be particularly well-suited for replicating intelligent behavior in low-power neuromorphic hardware.

%% file: Chapters/Chapter0A.tex
\chapter{Appendix}\label{ch:appendix}

\section{Implementation}
The code is available at \texttt{https://github.com/wkvanderveen/maspro}.
It does not require any machine learning libraries (except basic PyTorch to initialize arrays on the GPU); all computations are explicitly implemented in pure multidimensional NumPy-like arrays.
The code contains a configuration file to set various options for the MFCC preprocessing and e-prop process, including the number of layers, whether to use an uni- or bidirectional network (a legacy option not treated in this report), and the window size in the MFCCs, for example.

\begin{table}[ht]
    \myfloatalign
    \begin{tabularx}{\textwidth}{rrr} \toprule
        \tableheadline{Mels} & \tableheadline{Hz}
        & \tableheadline{Filterbank} \\ \midrule
        0    & 0\phantom{.0} & 0 \\
        105  & 68.5   & 2 \\
	        210  & 143.7  & 4 \\
	        315  & 226.2  & 7 \\
        420  & 316.8  & 10 \\
        525  & 416.3  & 13 \\
        630  & 525.5  & 16 \\
        735  & 645.4  & 20 \\
        840  & 777\phantom{.0} & 24 \\
        945  & 921.5  & 29 \\
        1050 & 1080.1 & 34 \\
        1155 & 1254.4 & 40 \\
        1260 & 1445.4 & 46 \\
        1365 & 1655.3 & 53 \\
        1470 & 1885.7 & 60 \\
        1575 & 2138.6 & 68 \\
        1680 & 2416.3 & 77 \\
        1785 & 2721.2 & 87 \\
        1890 & 3055.9 & 97 \\
        1995 & 3423.3 & 109 \\
        2100 & 3826.7 & 122 \\
        2205 & 4269.5 & 136 \\
        2310 & 4755.7 & 152 \\
        2415 & 5289.4 & 169 \\
        2520 & 5875.3 & 188 \\
        2625 & 6518.6 & 209 \\
        2730 & 7224.8 & 231 \\
        2835 & 8000\phantom{.0} & 256 \\
		\bottomrule
    \end{tabularx}
    \caption[Filterbanks]{Conversion table between linearly spaced Mels and their corresponding frequencies and filterbank boundaries.}
    \label{tab:mels}
\end{table}

\begin{table}[ht]
    \myfloatalign
    \begin{tabularx}{\textwidth}{lll} \toprule
        \tableheadline{Symbol} & \tableheadline{Description}
        & \tableheadline{Value} \\ \midrule
        $\alpha$              & Activity leak               & 0.8 \\
        $\beta$               & Adaptivity                  & 0.184 \\
        $\rho$                & Adaptivity leak             & 0.975 \\
        $\kappa$              & Output decay                & 0.8 \\
        $\gamma$              & Pseudoderivative dampening  & 0.3 \\
        $v_\text{th}$         & Base threshold              & 0.95 \\
        $\delta t_\text{ref}$ & Refractory time             & 2 \\
        $\eta$                & Learning rate               & 0.01 \\
        $\beta_1$             & Adam momemtum factor 1      & 0.9 \\
        $\beta_2$             & Adam momemtum factor 2      & 0.999 \\
        $c_\text{reg}$        & Firing rate regularization  & 50 \\
        $c_\text{L2}$         & L2 regularization           & $10^{-5}$ \\
        $f^\text{target}$     & Target firing rate          & 0.01 \\
        $N$                   & Network size                & $800^*$ \\

		\bottomrule
    \end{tabularx}
    \caption[Hyperparameters]{The full list of hyperparameter values, used in all networks and neuron model types. \\\textsuperscript{*}This is the total number of neurons in multi-layer networks (400 per layer in 2-layer networks, and 266 in 3-layer networks).}
    \label{tab:hparams}
\end{table}

\begin{figure}[bth]
    \myfloatalign
    \subfloat[ALIF cross-entropy.]
    {\includegraphics[height=3.4cm, keepaspectratio]{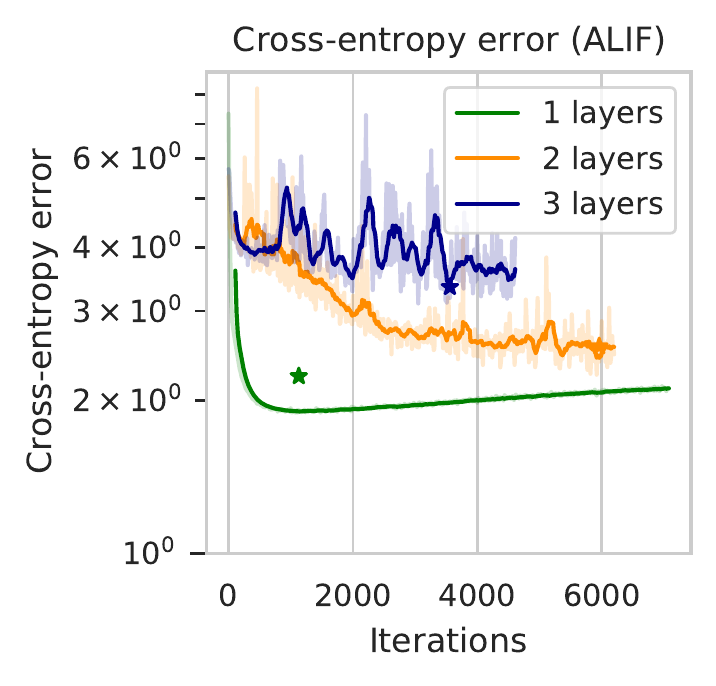}} \quad
    \subfloat[STDP-ALIF cross-entropy.]
    {\includegraphics[height=3.4cm, keepaspectratio]{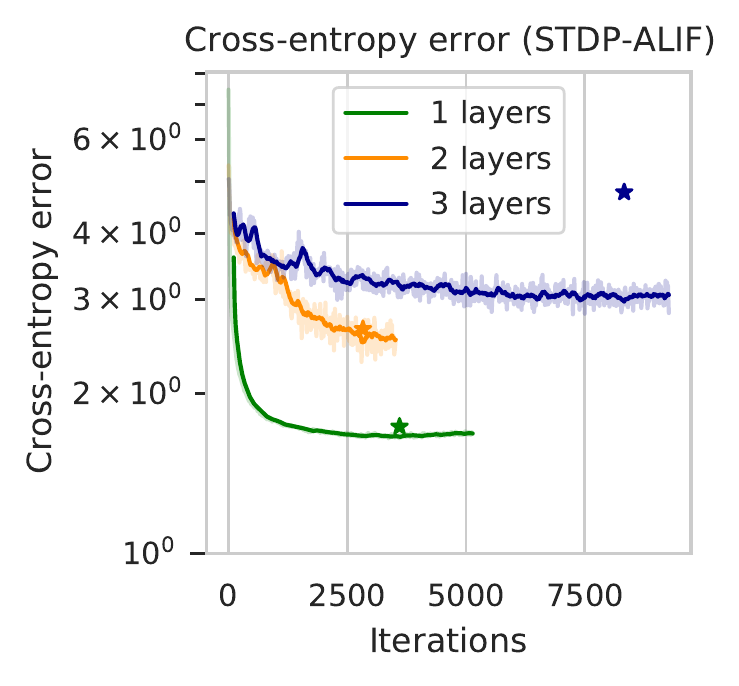}} \quad
    \subfloat[Izhikevich cross-entropy.]
    {\includegraphics[height=3.4cm, keepaspectratio]{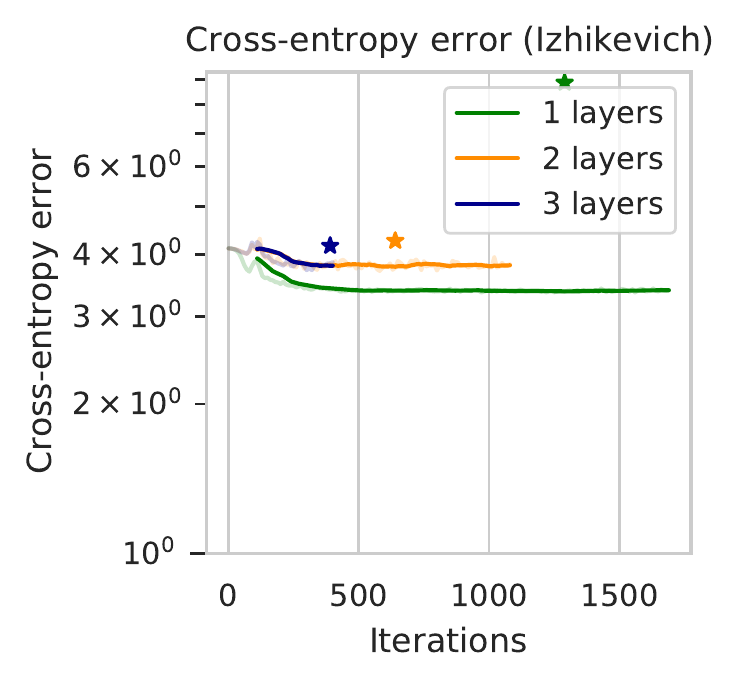}} \\
    \subfloat[ALIF spike rate.]
    {\includegraphics[height=3.6cm, keepaspectratio]{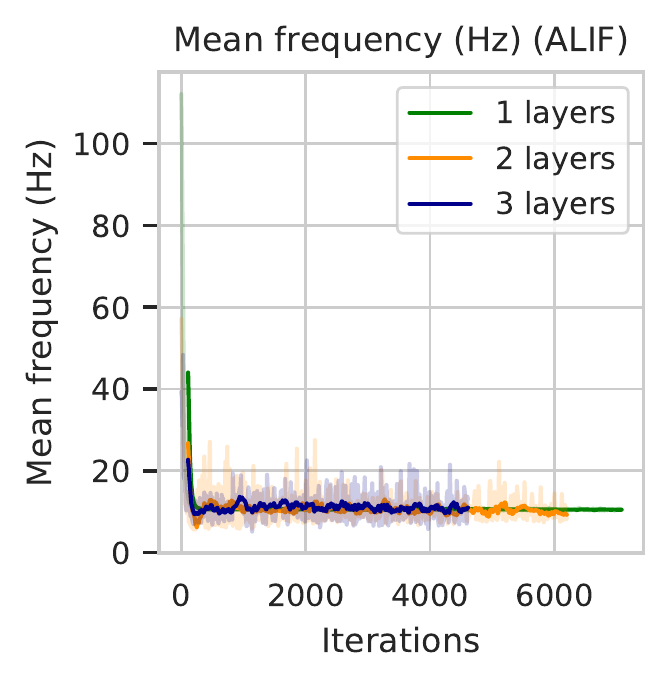}} \quad
    \subfloat[STDP-ALIF spike rate.]
    {\includegraphics[height=3.6cm, keepaspectratio]{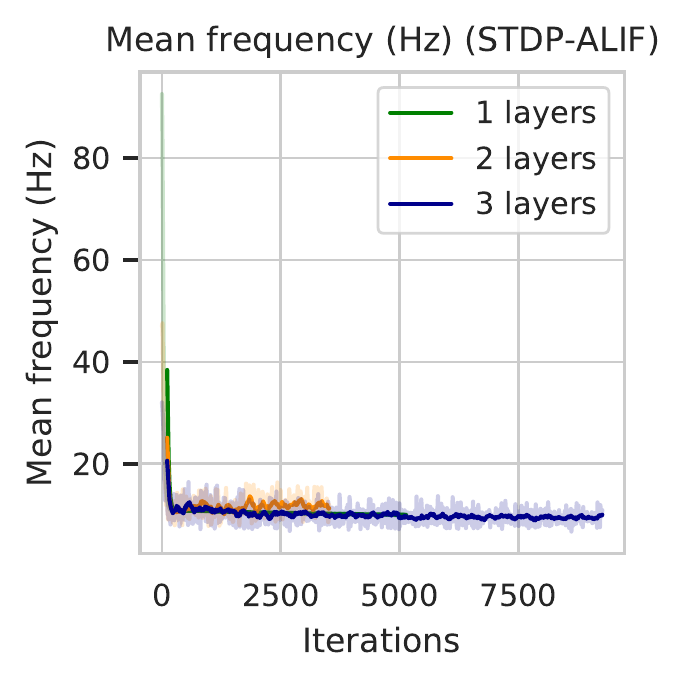}} \quad
    \subfloat[Izhikevich spike rate.]
    {\includegraphics[height=3.6cm, keepaspectratio]{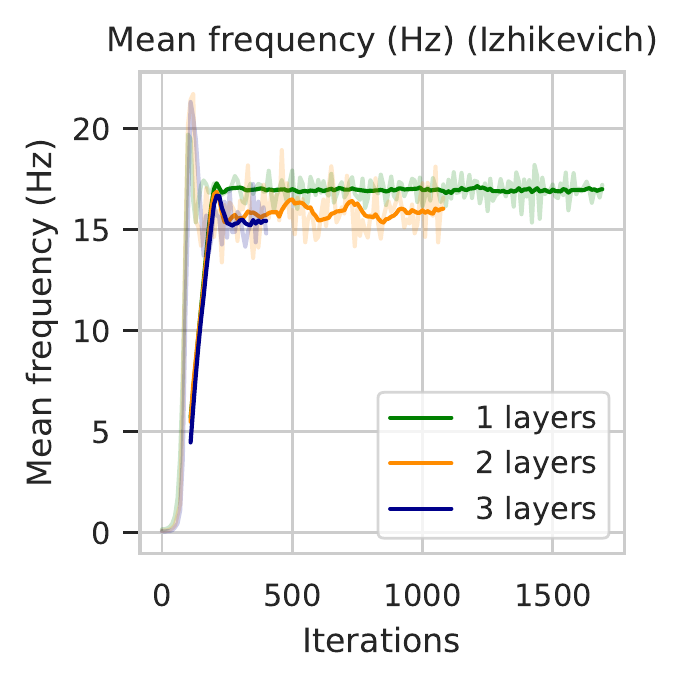}} \\
    \subfloat[ALIF reg. error.]
    {\includegraphics[height=3.6cm, keepaspectratio]{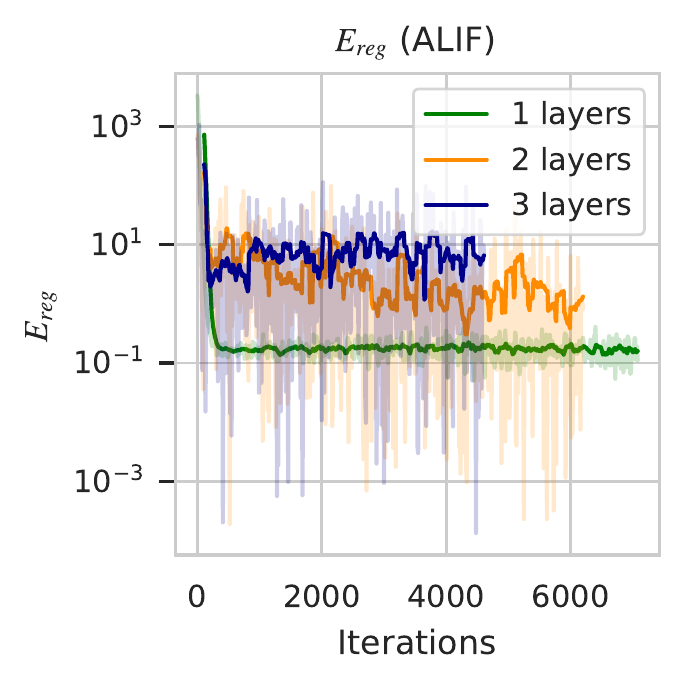}} \quad
    \subfloat[STDP-ALIF reg. error.]
    {\includegraphics[height=3.6cm, keepaspectratio]{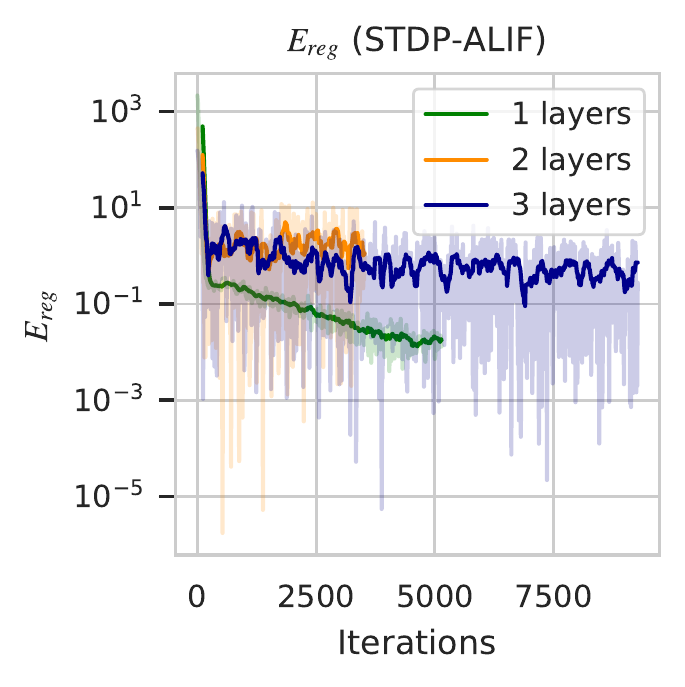}} \quad
    \subfloat[Izhikevich reg. error.]
    {\includegraphics[height=3.6cm, keepaspectratio]{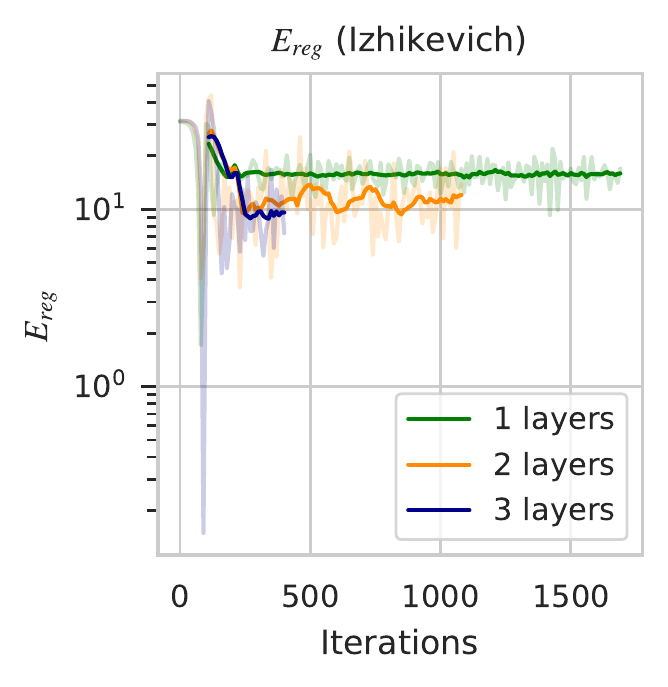}}
    \caption[Cross-entropy rates, mean spiking frequencies, and regularization errors for multi-layer networks.]{Cross-entropy rates, mean spiking frequencies, and regularization errors for multi-layer networks.}\label{fig:ml-otherresults}
\end{figure}

%% file: FrontBackmatter/Bibliography.tex
\defbibheading{bibintoc}[\bibname]{%
  \phantomsection
  \manualmark
  \markboth{\spacedlowsmallcaps{#1}}{\spacedlowsmallcaps{#1}}%
  \addtocontents{toc}{\protect\vspace{\beforebibskip}}%
  \addcontentsline{toc}{chapter}{\tocEntry{#1}}%
  \chapter*{#1}%
}
\printbibliography[heading=bibintoc]